\pdfoutput=1
\documentclass[11pt]{article}

\usepackage[preprint]{acl}

\usepackage{times}
\usepackage{latexsym}

\usepackage[T1]{fontenc}
\usepackage[utf8]{inputenc}

\usepackage{microtype}

\usepackage{inconsolata}

\usepackage{graphicx}

\usepackage{booktabs}   % 三线表样式
\usepackage{multirow} 
\usepackage{makecell} 
\usepackage{tabularx}
\usepackage{enumitem}
\usepackage{threeparttable}
\usepackage{adjustbox}       % 安全缩放支持
\usepackage{xurl} % 智能 URL 换行支持
\usepackage{hyperref}
\usepackage{amssymb}  % 提供 \checkmark 符号
\usepackage{amsmath}   % 基础数学符号支持
\usepackage{cuted}

\usepackage{tcolorbox}
\usepackage{multicol}
\tcbuselibrary{breakable} %breakable：支持跨页

\usepackage{cuted}
\title{Psychology-driven LLM Agents for Explainable Panic Prediction on Social Media during Sudden Disaster Events}

\author{
 \textbf{Mengzhu Liu\textsuperscript{*1}},
 \textbf{Zhengqiu Zhu\textsuperscript{*1}},
 \textbf{Chuan Ai\textsuperscript{1}},
 \textbf{Chen Gao\textsuperscript{2}},
  \\
 \textbf{Xinghong Li\textsuperscript{1}},
 \textbf{Lingnan He\textsuperscript{3}},
 \textbf{Kaisheng Lai\textsuperscript{4}},
 \textbf{Yingfeng Chen \textsuperscript{2}},
 \textbf{Xin Lu\textsuperscript{1}},
 \textbf{Yong Li\textsuperscript{2}},
 \textbf{Quanjun Yin\textsuperscript{1}}
\\
 \textsuperscript{1}National University of Defense Technology,
 \textsuperscript{2}Tsinghua University, \\
 \textsuperscript{3}Sun Yat-sen University,
 \textsuperscript{4}Jinan University \\
 \textsuperscript{*}Equal contribution
\\
}

\begin{document}
%\maketitle

\twocolumn[{
\renewcommand\twocolumn[1][]{#1}
\maketitle
\begin{center}
    \captionsetup{type=figure}
    \includegraphics[width=\textwidth]{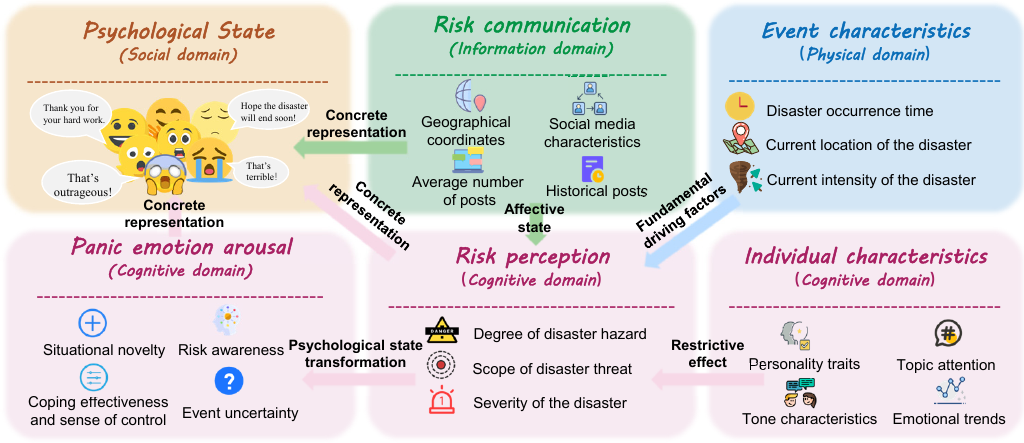}
    \captionof{figure}{Path of panic formation. The physical domain supplies an objective benchmark, while the information domain dynamically adjusts cognitive inputs. The cognitive domain processes these inputs through individual constraints and risk perception, triggering panic emotions that ultimately materialize as observable social media behaviors through the social domain.}
    \label{fig:path_of_panic_formation}
\end{center}
}]

\begin{abstract}
Accurately predicting public panic sentiment on social media is crucial for proactive governance and crisis management. Current efforts on this problem face three main challenges: lack of finely annotated data hinders emotion prediction studies, unmodeled risk perception causes prediction inaccuracies, and insufficient interpretability of panic formation mechanisms.
% (1) the lack of finely annotated panic emotion data hinders emotion prediction studies; (2) existing methods struggle to uniformly model the risk perception process involving multi-domain heterogeneous data interactions, causing prediction inaccuracies; (3) insufficient interpretability of panic formation mechanisms. 
We address these issues by proposing a \textbf{Psycho}logy-driven generative \textbf{Agent} framework (PsychoAgent) for explainable panic prediction based on emotion arousal theory. Specifically, we first construct a fine-grained panic emotion dataset (namely COPE) via human-large language models (LLMs) collaboration, combining scalable LLM-based labeling with human annotators for accuracy and bias mitigation. Then, we construct PsychoAgent integrating cross-domain heterogeneous data grounded in psychological mechanisms to model risk perception and cognitive differences in emotion generation. To enhance interpretability, we design an LLM-based role-playing agent that simulates \textit{individual psychological chains} through dedicatedly designed prompts.
% Finally, we implement a Mixture of Experts (MoE) system to ensure multi-dimensional consistency in generated responses, thereby boosting prediction accuracy. 
Experimental results on our annotated dataset show that PsychoAgent improves panic emotion prediction performance by 12.6\% to 21.7\% compared to baseline models. Furthermore, the explainability and generalization of our approach is validated.
% Ablation studies validate the effectiveness of each design component in emotion prediction tasks, while further experiments have demonstrated the explainability and generalization of our approach. 
Crucially, this represents a paradigm shift from opaque "data-driven fitting" to transparent "role-based simulation with mechanistic interpretation" for panic emotion prediction during emergencies. Our implementation is publicly available at: \url{https://anonymous.4open.science/r/PsychoAgent-19DD}.
\end{abstract}

% \maketitle
\section{Introduction}
Social media accelerates the spread of extreme emotions during sudden events \cite{world2020managing}, yet its real-time nature, rich information content, and spatiotemporal granularity offer unique opportunities for emotion prediction and management. Among various negative emotions, uncontrolled panic can exhaust emergency resources, trigger secondary disasters, even escalate isolated hazards into complex societal crises \cite{rune2023time}. For instance, during the 2011 Japan earthquake, panic-driven mass salt-buying in China exacerbated price surges and supply shortages \cite{pierson2011japan}. Hence, timely prediction and monitoring of panic enable proactive interventions to mitigate its spread and impact \cite{wu2025analyzing}.

Current research focuses on coarse-grained retrospective emotion classification using single-modality textual data \cite{mitrovic2024comparing,weigang2024evolution,mitrovic2022enhancing}, while dedicated studies on panic emotion prediction remain scarce. More details about emotion prediction research are in Appendix ~\ref{subsec:current_emotion_prediction}. Due to the limitations of dataset annotation quality and inherent methodological constraints, precisely predicting public panic remains significant challenges. 

Firstly, \textbf{current panic emotion annotation methods have semantic deviations.} Existing datasets rely on static annotations and lack normal-state behavioral data, impeding predictive modeling of panic evolution and causing: (1) oversimplified categorization of panic as fear \cite{yang2022spatial}, ignoring disaster-specific risk perception; (2) semantic drift in keyword-based detection due to variations in linguistic nuances and expressive conventions \cite{mitrovic2024comparing,weigang2024evolution}, weakening cognition-semantics alignment.

Secondly, \textbf{current prediction models have cross-domain feature fusion and dynamic modeling limitations.} Psychological studies indicate that public emotions are influenced by multi-domain heterogeneous features. While current models attempt to incorporate such features \cite{regan2024can}, they remain limited to simplistic combinations of text with single factors. Moreover, existing models often use static temporal modeling, fail to capture evolving feature interactions, significantly reduce predictive efficacy \cite{mitrovic2024comparing}.

Lastly, \textbf{current studies have explainability gap in panic formation mechanism.} Current studies focus on surface emotion detection and statistical correlations \cite{mitrovic2024comparing}, neglecting psychological drivers like risk perception, cognitive appraisal mechanisms \cite{HariharanDWA17}. Though cognitive-emotion interactions are validated in psychological studies, existing works fail to computationally map the risk perception-to-arousal chain \cite{regan2024can,houlihan2023emotion}, leaving emotional evolution pathways unexplained.

To address the above challenges, we propose a \textbf{Psycho}logy-driven generative \textbf{Agent} framework (PsychoAgent) for panic emotion prediction during sudden disasters. Specifically, we first adopt a human-LLM collaborative annotation method to construct a fine-grained panic emotion dataset (\textbf{C}ollaborative Fine-grained \textbf{O}pen \textbf{P}anic \textbf{E}motions Dataset, COPE), resolving semantic deviation and cross-cultural annotation biases. Secondly, we develop a mental framework to fuse multi-domain features through the psychological mechanisms of panic formation (as illustrated in Figure ~\ref{fig:path_of_panic_formation}). Finally, guided by theoretical lenses of risk perception and emotion arousal, we design a chain-of-thought (CoT)-driven LLM-based agent to simulate the full psychological chain of "disaster perception, risk cognition, panic emotion arousal, and posting behavior response". Our work shows significant implications for developing explainable AI systems capable of providing deeper insights into collective emotional dynamics during sudden disaster events.

Our contributions are summarized as:
\begin{itemize}[leftmargin=*]
\vspace{-6pt}
\item To support panic prediction, we pioneer a novel dual-phase panic emotion annotation dataset. We also fine-tune a discriminator to detect panic signals in generated texts.
\vspace{-9pt}
\item We innovatively propose a psychology-driven, multi-domain fusion mental model for public panic prediction during disasters. Our approach integrates multi-domain data via an LLM-based role-playing agent for interpretable psychological modeling of panic formation mechanisms.
\vspace{-22pt}
\item Experimental results show that the proposed framework achieves an accuracy of $\mathbf{86\%}$ ($\uparrow$ at least $\mathbf{13\%}$ versus SOTA) in panic emotion prediction task. Moreover, scalability studies and case studies further validate the explainability and generalization of our approach.
\end{itemize}

\section{Problem Definition and Dataset Construction}

In this section, we first formalize the problem definition, then detail the methods for building a fine-grained panic emotion dataset and finally fine-tune a BERT-based model for panic recognition.

% In this section: 
% \begin{itemize}[itemsep=2pt, topsep=0pt, parsep=0pt ]
%     \item we first formalize the problem definition of our study in Section 2.1.
%     \item Then in Section 2.2, we describe the methodology for constructing the fine-grained annotated panic emotion dataset and demonstrate the effectiveness of fine-tuning a BERT model for panic emotion recognition.
%     \item In Section 3.3, we elaborate on our proposed PsychoAgent panic prediction framework.
%     \item In Section 3.4, we detail the extraction of individual characteristics.
%     \item  And in Section 3.5, we explain the method for CoT-driven LLM-based agent generation of predictive text responses.
% \end{itemize}

\subsection{Problem Definition}
%Current retrospective sentiment analysis tasks can be formulated as follows:
%\begin{equation}
%\sigma_t^i=F\left( I_t^i\right)
%\end{equation}
%where ${\sigma_t^i}$ is the sentiment label of user $i$ at time $t$, and $I_t^i$ is post contents that user $i$ publishes on social media at time $t$. 

The sentiment prediction task can be defined as:
\begin{equation}
\sigma_{t+1}^i=F\left( I_t^i\right),
\end{equation}
where $\sigma_{t + 1}^{i}$ is the sentiment label of user $i$ at future time $t + 1$, and $I_t^i$ is post contents that user $i$ publishes on social media at time $t$. 

This study aims to predict users' panic emotion labels after disaster occurrence through deep fusion of multi-domain features from their pre-disaster social media posts, integrated with psychological theories. Formally, this task can be formulated as:
\begin{equation}
\sigma_{t+1}^{i} = F\left( I_{t}^{i},P_{t+1}, C_{t}^{i} \right),
\end{equation}
where $t+1$ is the time after the sudden disaster, $t$ is the time before the disaster, $\sigma_{t+1}^i$ is the post-disaster sentiment label of user $i$, $P_{t+1}$ is the post-disaster physical features, and $C_t^i$ is the personal traits of user $i$ from pre-disaster analysis.

\subsection{COPE Dataset Benchmark}
To address the gap in high-quality panic emotion resources, we develop the first fine-grained dataset COPE, spanning pre- and post-disaster phases, which fills the gap in standardized panic emotion datasets and provides a foundation for advancing research in emotion prediction and related fields. Its core innovation is the standardized human-LLM collaborative annotation process, structured as follows. More details are in Appendix ~\ref{subsec:Dataset_BERT}. 
\begin{itemize}[leftmargin=*]
    \vspace{-8pt}
    \item \textbf{Multi-source data collection: }Focusing on Hurricane Sandy (2012), we used 52.25 million tweets from 13.75 million users (Oct. 15- Nov. 12) \cite{kryvasheyeu2015performance}. After preprocessing (detailed in Appendix ~\ref{subsubsec:Data_preprocessing}), the dataset includes 9,065 users, including user IDs, follower/followee counts, timestamps, locations, and ternary emotion labels. 
    \vspace{-9pt}
    \item \textbf{Panic annotation framework: }Targeting post-disaster texts, we adopt a LLM-human hybrid annotation pipeline: (1) LLM generates initial labels via semantic parsing; (2) 100 annotators conducted three rounds of cross-validation to ensure robustness; (3) Fine-tuned BERT model automates remaining annotations.
    \vspace{-10pt}
    \item \textbf{User-level panic identification: }We identify panicked users via a "one-veto" rule ($\geq$1 panic text).
\end{itemize}

\section{Psychology-Driven LLM Agent Framework}
We first delineate the architecture of PsychoAgent. Then we describe the methodology for individual feature extraction. Finally, we introduce the psychology-driven LLM-based agent to simulate posting behavior for predicting panic emotions.
% \begin{itemize}[itemsep=2pt, topsep=0pt, parsep=0pt ]
% \item we first present the proposed PsycoAgent framework in Section 3.1;
% \item Then in Section 3.2, we describe the details of individual feature extraction;
% \item  In Section 3.3, we introduce the CoT-driven LLM-based agent to generate predictive textual responses.
% \end{itemize}

\subsection{Overview of PsycoAgent}
Based on the studies of risk perception and emotion arousal in psychology, we summarize the panic emotion formation pathway as follows (Figure \ref{fig:path_of_panic_formation}): 

\begin{itemize}[leftmargin=*]
\vspace{-7pt}
\item In the physical domain, we transform spatiotemporal disaster intensity into interpretable data through risk communication;
\vspace{-10pt}
\item In the information domain, we amplify disaster severity through communication channels, modulating risk perception and refining physical data interpretation, establishing via physical-cognitive bidirectional feedback loops;
\vspace{-10pt}
\item In the cognitive domain, we utilize individual traits to mediate risk interpretation biases, triggering panic through emotional arousal to achieve the transformation from data to emotion;
\vspace{-10pt}
\item In the social domain, we materialize individual panic into collective psychological states through social media behavior patterns.
\end{itemize}

\begin{figure*}[ht]
  \centering
  \includegraphics[width=\textwidth]{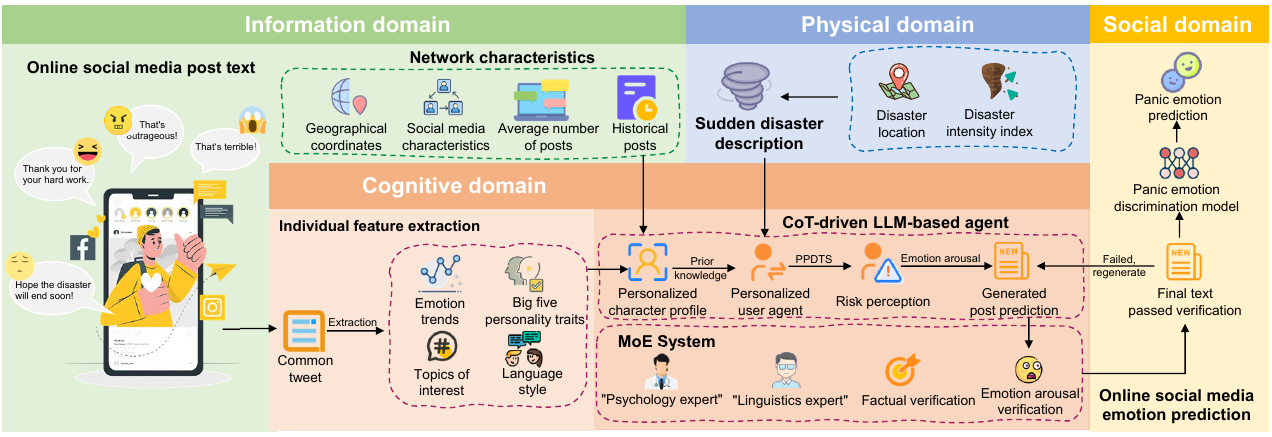}
  \caption{Overview of the proposed panic emotion prediction framework——PsychoAgent.}
  \label{fig:framework}
\end{figure*}

\vspace{-7pt}
Based on this, we propose PsychoAgent (Figure ~\ref{fig:framework}), a multi-domain fusion-driven LLM agent framework for panic emotion prediction, which comprises four main parts: \textbf{(1) In the information domain}, we focus on collecting social media texts and risk communication channel features; \textbf{(2) In the physical domain}, we extract disaster characteristics via geospatial platforms, gathering key indicators to convert physical features into disaster information; \textbf{(3) In the cognitive domain}, we extract individual traits from pre-disaster social media posts. Then, we construct user profiles to drive the LLM agents, aiming at simulating "disaster perception, risk perception, panic arousal, and posting behavior" psychological chain; \textbf{(4) In the social domain}, we use the fine-tuned BERT model to verify the generated text. Then, we enforce a "one-vote" veto rule for user-level prediction. Subsequent sections detail cognitive domain implementations.

\subsection{Psychological and Behavioral Features Extraction}
Psychological studies establish that personality traits, sentiment dynamics, topical concerns, and linguistic patterns critically shape risk perception and emotional responses \cite{gross2003individual,brosch2013impact,syrdal2018engagement,davis2020transformation}. We accordingly extract four main features from users' pre-disaster posts.

\paragraph{Big Five Personality Traits}
We employ the publicly available model bert-base-personality \cite{devlin2019bert} to analyze the user's personality traits. This process can be formally expressed as:
\begin{equation}
\rho^{i} = \psi_{\text{personality}}( I_{t}^{i}),
\end{equation}
where $\rho^i$ is the personality vector of the $i$-th user, and $\psi_{\text{personality}}$ is the model. More details are provided in Appendix~\ref{subsubsec:big_five_traits}.

\paragraph{Sentiment Trend Features}
We use the BERTweet model \cite{nguyen2020bertweet} to perform ternary sentiment analysis on pre-disaster posts, aggregating results to generate an emotional trend profile. This is formalized as:
\begin{equation}
\gamma^{i} = \psi_{\text{sentiment}} (I_{t}^{i}),
\end{equation}
where $\gamma^{i}$ is the sentiment trend vector of the $i$-th user; $\psi_{\text{sentiment}}$ is the sentiment model.

\paragraph{Topic-Aware Characteristic Features}
Using LDA model clustering and LLM-guided merging, we extract topic features by analyzing users' historical posts. The process is expressed as:
\begin{equation}
\tau^i=\Gamma\cdot\Theta^i,
\end{equation}
where $\Theta^{i}$ is the LDA-clustered topic vector of the $i$-th user; $\Gamma$ is the membership relationships; $\tau^{i}$ is the consolidated thematic focus vector summarized by LLM. Appendix~\ref{subsubsec:topic_feature} offers additional details.

\paragraph{Linguistic Features}
We design prompts to capture users' tone features via LLM. The process is formalized as:
\begin{equation}
v^{i} = \text{LLM}_{\text{language}}(I_{t}^{i}, p_{v}),
\end{equation}
where $v^{i}$ is the tone features of the $i$-th user; $\text{LLM}_{\text{language}}$ is the tone feature extraction model; $p_{\nu}$ is the prompt for LLM to extract tone features. More details can be referred to Appendix~\ref{subsubsec:Tone_feature}.

\subsection{Building LLM Agent with Psychological Theories}
\begin{figure*}[ht]
  \centering
  \includegraphics[width=\linewidth]{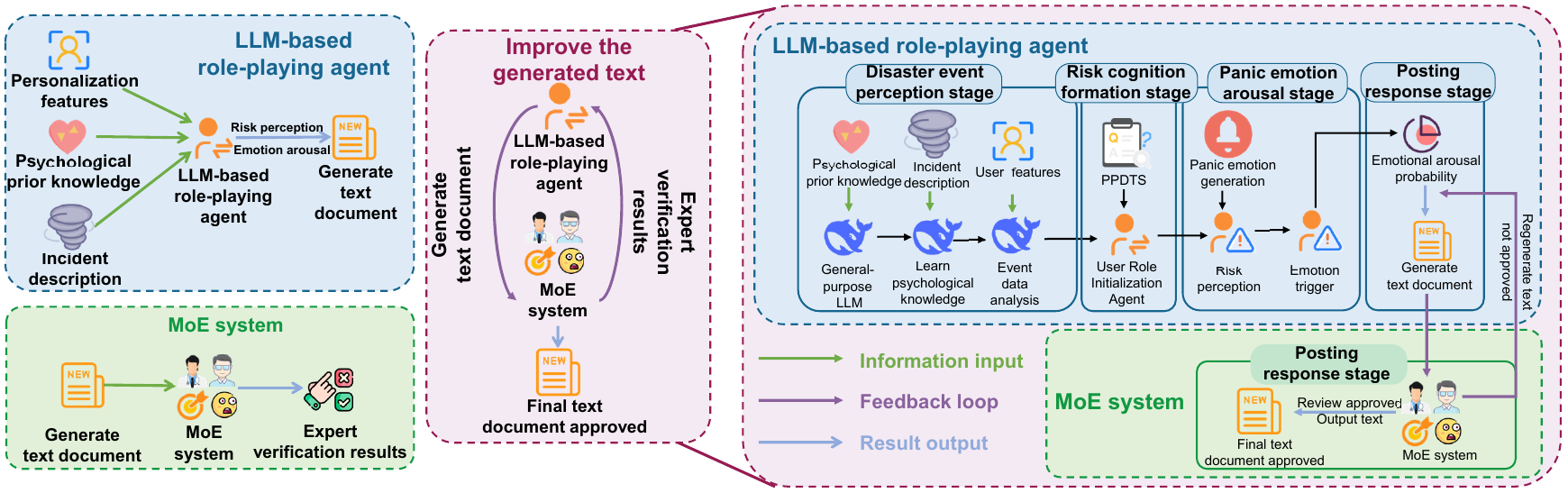}
  \caption{Detailed stages for building LLM-based agent driven by CoT in the cognitive domain module.}
  \label{fig:llm_agent_steps}
\end{figure*}

% Generative agents are LLM-based systems that simulate human behavior via enhanced memory \cite{park2023generative}. 
We design a CoT-driven LLM-based agent to simulate users' perception and responsiveness through four stages, with key stages centering on risk perception and emotional arousal (see in Figure \ref{fig:llm_agent_steps}):

\begin{itemize}[leftmargin=*]
  \vspace{-10pt}
  \item Disaster perception stage (physical-information domain): We integrate psychological knowledge, disaster features, and user profiles to form agent's long-term memory;
  \vspace{-10pt}
  \item Risk perception stage (cognitive domain): We prompt the agent to engage with the psychological preparedness for disaster threats scale (PPDTS) \cite{mclennan2020conceptualising};
  \vspace{-10pt}
  \item Panic arousal stage (cognitive domain): We prompt the agent to quantify panic probability through multi-dimensional analysis;
  \vspace{-10pt}
  \item Posting response stage (social domain): We generate tweets with MoE-based consistency verification mechanism.
  % \item Panic prediction stage: We use a "one-veto" rule to predict user panic states based on validated generated texts.
\end{itemize}

\vspace{-6pt}
\paragraph{Disaster Perception Stage}
This stage equips the agent with essential prior knowledge by integrating psychological knowledge, disaster data, and user features into its long-term memory via structured prompts. The process is expressed as:

\begin{equation}
\text{LLM}_U^i=\text{LLM}_\text{RP}(K,D,U^i,p_u),
\end{equation}
where $\text{LLM}_U^i$ denotes the user feature learning agent for user $i$; $\text{LLM}_\text{RP}$ is the role-playing LLM-based agent; $K$ is psychological knowledge constraints; $D$ is disaster data; $U^{i}$ is features of user $i$, encompassing both static traits and risk communication patterns; $p_{u}$ is the initialization prompt. Detailed implementation is in Appendix ~\ref{subsubsec:psycho_konwledge}.

\paragraph{Risk Perception Stage}
We construct a dynamic risk perception framework using disaster perception stage's memory and PPDTS (see Appendix~\ref{subsubsec:PPTDS}), emulating users' risk perception processes via structured psychological assessment. By leveraging LLM's reasoning capabilities with a human cognitive-inspired stepwise CoT prompting mechanism: question comprehension, memory retrieval, option mapping, and summary judgment, we prompt the agent to output risk perception scores (1-4 levels) for each question, formalized as: 
\begin{equation}
S^i=\text{LLM}_U^i(Q,p_{\text{perception}}),
\end{equation}
where $S^{i}$ is the PPDTS evaluation scores for user $i$; $Q$ is the question set comprising the PPDTS; $ p_{\text{perception}}$ is the PPDTS assessment prompt.

\paragraph{Panic Arousal Stage}
To bridge risk perception and emotion arousal, we reparameterize the discrete PPDTS scores into panic propensity drivers. Specifically, we prompt the agent to analyze four core factors (Risk awareness, coping effectiveness, event uncertainty, and situational novelty) through a cognitive chain reasoning mechanism, and infer the panic propensity probability based on prior outputs. This process can be represented as:

\begin{equation}
C^i=\text{LLM}_U^i(S^i,p_c),
\end{equation}
\begin{equation}
P_{\text{panic}}^i=\text{LLM}_U^i(S^i,C^i,p_{\text{panic}}),
\end{equation}
where $C^{i}$ is the summary of four factors for user $i$; $P_{\text{panic}}^i$ is the panic propensity probability; $P_C$ is the prompt directing the agent in summarizing the four factors; $p_{\text{panic}}$ is the prompt guiding the agent to evaluate user's panic probability.

\paragraph{Posting Response Stage}
In this stage, we prompt the agent to integrate knowledge across stages (see Appendix ~\ref{subsubsec:MoE_system} for details) and predict the post content the user is most likely to generate under the disaster event. This process can be represented as:

\begin{equation}
T^i=\text{LLM}_U^i(S^i,C^i,P_{\text{panic}}^i,p_T),
\end{equation}
where $T^i$ is the set of tweets most likely published by the $i$-th user; $P_{T}$ is the prompt guiding the agent to generate predictive texts. 

To mitigate subjective bias and randomness risks in evaluation, we introduce a MoE system with psychological, linguistic, factual consistency, and emotional alignment models, which assesses the generated text's multi-dimensional consistency. The details are provided in Appendix ~\ref{subsubsec:MoE_system}. This process can be represented as: 
\begin{equation}
    \varepsilon_k^i=\text{LLM}^K(U^i,D,P_{\text{panic}}^i,T^i,p_k),
\end{equation}
%\begin{equation}
    %\partial=\bigcap_{k=1}^4\mathcal{E}_k,
%\end{equation}
where $\varepsilon_k^i \in {0,1}$ is the evaluation result of the $k$-th expert for the $i$-th user’s generated text; $\text{LLM}^K$ is the $k$-th expert model induced via prompt $p_{k}$. 

The final evaluation result is determined by all expert models. The generated text $T^{i}$ is only validated if all expert models give a positive evaluation. Otherwise, the system adjusts and retries based on the experts' feedback. Further details of prompts and case demonstrations are in Appendix ~\ref{subsubsec:case_demonstrations}.

%\begin{equation}
%T_g^i=LLM_U^i(\varepsilon_k^i,S^i,C^i,P_{panic}^i,p_g)
%\end{equation}

%where $T_{g}^{i}$ is the re-generated predicted text for the $i$-th user; $p_{g}$ is the prompt for re-generating the predicted text.

% \subsubsection{Panic Emotion Prediction}
% Finally, we input the validated generated predictive texts into the self-trained panic emotion detection BERT model to identify panic emotion labels. We strictly enforce a "one-vote" veto rule for user-level prediction: if any of the three generated texts for a user is labeled as panic emotion data, the user is classified as a panic user. This process ensures high sensitivity and accuracy in emotional state assessment under complex scenarios.

\section{Experiment}
In this section, we conduct extensive experiments on COPE to validate the proposed framework, including performance comparison, ablation study, scalability study and case study, with more experimental details provided in Appendix~\ref{subsec:baseline} and~\ref{subsec:case_studies}.
% In this section, we conduct systematic experiments on our self-developed fine-grained panic emotion annotated dataset to validate the proposed framework through empirical analysis, addressing four research questions:

% \begin{itemize}
 % \item \textbf{RQ1 (Performance Validation):} How does our multi-domain fusion framework grounded in emotion arousal theory compare to SOTA emotion prediction baselines in terms of prediction accuracy?
 % \item \textbf{RQ2 (Module Ablation):} Does the CoT-driven generative agent function as intended across reasoning phases? How does inter-module synergy affect overall performance?
 % \item \textbf{RQ3 (Model Scalability):} How does the framework perform when using LLMs of varying parameter sizes? Can smaller models achieve comparable performance to large ones in resource-constrained settings?
 % \item \textbf{RQ4 (Case Analysis):} How does the proposed method demonstrate superior mechanistic interpretability compared to traditional data-fitting approaches?
% \end{itemize}

% In what follows, we first describe the experimental settings, and then answer the above four research questions.

\subsection{Experimental Setup}
\paragraph{Evaluation Metrics}
To address the class imbalance in panic detection, we adopt five macro-averaged metrics (accuracy, AUC, precision, recall, F1-score) \cite{sokolova2009systematic}, which compute scores per class and average them to mitigate majority-class bias.
%\begin{itemize}
    %\item Accuracy: Overall prediction correctness;
    %\item AUC: Ability to distinguish positive/negative classes;
    %\item Macro Precision: Ratio of true positives among predicted positives, reflecting false-positive suppression;
    %\item Macro Recall: Ratio of true positives correctly identified, indicating coverage completeness;
    %\item Macro F1-score: Harmonic mean of precision and recall, balancing both aspects.
%\end{itemize}

\begin{table*}[ht]
  \centering
  \begin{threeparttable}
  \scriptsize % 将字体设置为更小尺寸
  \renewcommand{\arraystretch}{0.6} % 减小行间距
  \setlength{\tabcolsep}{9pt} % 增加列间距
  \begin{tabular}{llccccccc}
    \toprule
    \textbf{Category} & \textbf{Method} & \textbf{} & \textbf{Accuracy} & \textbf{AUC} & \textbf{Precision} & \textbf{Recall} & \textbf{F1-score}  & \textbf{Support} \\
    \midrule
    
    % ============ Traditional ML Methods ============
    \multirow{12}{*}{\textbf{Traditional ML Methods}}
    & \multirow{3}{*}{SVM} 
    & Panic      & -  & - & 0.56 & 0.70 & 0.62 & 630 \\
    & & No Panic  & -  & - & 0.82 & 0.70 & 0.76 & 1183 \\
    & & Average  & 0.70 & 0.76 & 0.69 & 0.70 & 0.69 & 1813 \\
    \cmidrule(lr){2-9}
    
    & \multirow{3}{*}{Logistic Regression}
    & Panic     & -  & - & 0.55 & 0.73 & 0.63 & 630 \\
    & & No Panic & -  & - & 0.83 & 0.68 & 0.74 & 1183 \\
    & & Average & 0.70 & 0.75 & 0.69 & 0.70 & 0.69 & 1813 \\
    \cmidrule(lr){2-9}
    
    & \multirow{3}{*}{Random Forest}
    & Panic      & -  & - & 0.61 & 0.62 & 0.62 & 630 \\
    & & No Panic  & -  & - & 0.80 & 0.79 & 0.79 & 1183 \\
    & & Average   & \underline{0.73} & \underline{0.80} & \underline{0.70} & 0.71 & \underline{0.70} & 1813 \\
    \cmidrule(lr){2-9}
    
    & \multirow{3}{*}{XGBoost}
    & Panic      & -  & - & 0.60 & 0.61 & 0.61 & 630 \\
    & & No Panic & -  & - & 0.79 & 0.78 & 0.79 & 1183 \\
    & & Average  & \underline{0.73} & 0.79 & \underline{0.70} & 0.70 & \underline{0.70} & 1813 \\
    \midrule
    
    % ============ Deep Learning Methods ============
    \multirow{9}{*}{\textbf{Deep Learning Methods}}
    & \multirow{3}{*}{Bi-LSTM}
    & Panic      & -  & - & 0.56 & 0.60 & 0.58 & 648 \\
    & & No Panic & -  & - & 0.77 & 0.73 & 0.75 & 1165 \\
    & & Average  & 0.69 & 0.74 & 0.66 & 0.67 & 0.67 & 1813 \\
    \cmidrule(lr){2-9}
    
    & \multirow{3}{*}{Transformer}
    & Panic     & -  & - & 0.58 & 0.73 & 0.65 & 648 \\
    & & No Panic & -  & - & 0.82 & 0.71 & 0.76 & 1165 \\
    & & Average & 0.71 & 0.71 & \underline{0.70} & \underline{0.72} & \underline{0.70} & 1813 \\
    \cmidrule(lr){2-9}
    
    & \multirow{3}{*}{PredNet}
    & Panic      & -  & - & 0.56 & 0.59 & 0.57 & 648 \\
    & & No Panic  & -  & - & 0.76 & 0.74 & 0.75 & 1165 \\
    & & Average  & 0.69 & 0.68 & 0.66 & 0.67 & 0.66 & 1813 \\
    \midrule
    
    % ============ LLM-based Methods ============
    \multirow{9}{*}{\textbf{LLM-based Methods}}
    & \multirow{3}{*}{Deepseek-v3}
    & Panic      & -  & - & 0.50 & 0.00 & 0.01 & 562 \\
    & & No Panic & -  & - & 0.65 & 1.00 & 0.79 & 1065 \\
    & & Average  & 0.65 & 0.50 & 0.60 & 0.65 & 0.52 & 1627 \\
    \cmidrule(lr){2-9}
    
    & \multirow{3}{*}{ChatGPT-4o}
    & Panic      & -  & - & 0.67 & 0.01 & 0.02 & 630 \\
    & & No Panic & -  & - & 0.65 & 1.00 & 0.79 & 1183 \\
    & & Average  & 0.65 & 0.50 & 0.66 & 0.50 & 0.40 & 1813 \\
    \cmidrule(lr){2-9}
    
    & \multirow{3}{*}{THOR-ISA}
    & Panic      & -  & - & 0.00 & 0.00 & 0.00 & 630 \\
    & & No Panic  & -  & - & 0.65 & 1.00 & 0.79 & 1183 \\
    & & Average  & 0.65 & 0.50 & 0.33 & 0.50 & 0.39 & 1813 \\
    \midrule
    
    % ============ Our Method ============
    \multirow{3}{*}{\textbf{Our Method}}
    & \multirow{3}{*}{PsychoAgent}
    & Panic      & -  & - & 0.74 & 0.90 & 0.81 & 581 \\
    & & No Panic & -  & - & 0.94 & 0.84 & 0.89 & 1130 \\
    & & Average  & \textbf{0.86} & \textbf{0.87} & \textbf{0.84} & \textbf{0.87} & \textbf{0.85} & 1711 \\
    
    \bottomrule
  \end{tabular}

  \begin{tablenotes}[para]
    \scriptsize
    \raggedright
    \item[] \textbf{Note:} 
    (a) The Deepseek-v3 method ultimately predicts only 1,627 users due to sensitive content in user historical texts;
    (b) PsychoAgent successfully predicts 1,711 users after Stage 4 questionnaire validity screening, excluding cases with invalid questionnaires containing fewer than 18 responses (invalidity criterion: returned questions $<$ 18).
  \end{tablenotes}
  \end{threeparttable}
  \caption{The panic emotion prediction performance of PsychoAgent and the baselines, where the best performance is shown in \textbf{bold} and the second best is \underline{underlined}.}
  \label{tab:results}
\end{table*}

\paragraph{Implementation Details}
The experiments are driven by the DeepSeek-v3. During risk perception and emotion arousal analysis, the temperature is set to 0.4 for stable psychological reasoning; In tweet generation, it rise to 0.7 (with a repetition penalty of 0.4) for linguistic diversity; In text verification, the temperature reset to 0.4 for rigorous checks, with 3 adaptive retries. For baselines, we adapt models to the psychology prediction task while adhering to original technical specifications. All models use the same input features, data split (8:2), and evaluation metrics to ensure comparability.

\paragraph{Baselines}
Although research on panic emotion prediction remains exploratory, affective computing offers a robust methodological framework with extensive prior work. To validate our framework’s effectiveness, we choose three mainstream types of emotion analysis and prediction methods. Details of these baselines can be found in Appendix ~\ref{subsec:baseline}.

\begin{itemize}[leftmargin=*]
    \vspace{-7pt}
    \item Feature-engineered ML (machine learning) models: We select classic ML methods like SVM \cite{cortes1995support}, Logistic Regression \cite{hosmer2013applied}, Random Forest \cite{breiman2001random}, and XGBoost \cite{chen2016xgboost} as baselines. For each, we integrate user features including big five personality traits, sentiment trend features, linguistic features, and topic-aware characteristic features.
    \vspace{-7pt}
    \item End-to-end DL (deep learning) models: We select classic DL models like Bi-LSTM \cite{graves2005framewise}, Transformer \cite{vaswani2017attention}, and PredNet \cite{lotter2016deep} as baselines, with inputs similar to ML models.
    \vspace{-7pt}
    \item Advanced pre-trained language models: We use few-shot prompting on DeepSeek-v3 \cite{liu2024deepseek} and ChatGPT-4o \cite{achiam2023gpt} to directly generate panic prediction results as baselines. We also evaluate the THOR-ISA \cite{fei2023reasoning} three-step prompting framework for panic prediction and include it as a baseline.
\end{itemize}

%\subsection{Results}
\subsection{Performance Comparison with Baselines}
%To compare the predictive accuracy of our PsycoAgent framework with baseline methods, we conducted extensive  performance experiments. 
As shown in Table~\ref{tab:results}, our approach significantly outperforms the SOTA baselines across all evaluation metrics, demonstrating the following strengths:

\textbf{Psychological theory-driven framework design: }The PsychoAgent framework achieves an absolutely $\mathbf{7\%}$ AUC improvement (0.87 vs 0.80 for Random Forest). Its core advantage lies in the task-specific design for emotion prediction, which integrates psychological priors to model cognitive-emotional chains. Unlike static feature-based ML models, PsychoAgent dynamically simulates user psychological states, enhancing minority-class detection under class imbalance.

\textbf{Temporal dynamic modeling of multi-domain fusion:} The PsychoAgent framework demonstrates superior temporal dynamic modeling capabilities, surpasses Bi-LSTM (0.74), Transformer (0.71), and PredNet (0.68) by $\mathbf{12\%-18\%}$ AUC. The PsychoAgent overcoming Bi-LSTM's fixed time window limitations for abrupt psychological shifts and PredNet's static feature constraints which can't integrate physical and cognitive features effectively. Its phased memory-augmented architecture enables precise capture of both disaster dynamics and abrupt psychological mutations, validating robust multi-domain dynamic modeling.

\textbf{Limitations of LLMs and our strategy:} Although LLMs exhibit strong semantic understanding, their direct classification performance suffer from negative-class overprediction (precision: 0.66, recall: 0.50) due to mismatch between generative probability outputs and classification’s hard decision boundaries. PsychoAgent addresses this by decoupling LLM-based semantic parsing and risk simulation from direct label generation, achieving balanced precision (0.84) and recall (0.87), validating this indirect utilization strategy’s efficacy.

Overall, these results demonstrate the superior multi-domain dynamic modeling and minority class recognition capabilities of PsychoAgent. 

\subsection{Ablation Studies} 
In this section, we conduct ablation studies to analyze the contributions of key components in the PsychoAgent framework. By progressively removing key components: risk sensing (RS), emotion arousal (EA), and multi-expert assessment (MEA), we evaluate their individual and synergistic impacts on performance. The results are summarized in Table ~\ref{tab:ablation}, which demonstrate the full framework excels across all metrics. Specifically:
%Removing any key component significantly degrades performance, highlighting each component's vital role in the overall effectiveness. 

% 消融实验表格
\begin{table}[ht]
    \centering
    \scriptsize % 将字体设置为更小尺寸
    \renewcommand{\arraystretch}{0.4} % 减小行间距
    \newcolumntype{C}[1]{>{\centering\arraybackslash}p{#1}} % 自定义居中列类型
    \newcolumntype{R}[1]{>{\raggedleft\arraybackslash}p{#1}} % 自定义右对齐列类型
    \setlength{\tabcolsep}{1.0pt} % 压缩列间距
    \begin{tabular}{@{} p{2cm} >{\raggedright}p{1.0cm} C{0.7cm} C{0.7cm} C{0.7cm} C{0.7cm} C{0.7cm} C{0.7cm} @{}}
        \toprule
        \textbf{Method} & & \textbf{Acc.} & \textbf{AUC} & \textbf{Prec.} & \textbf{Recall} & \textbf{F1} & \textbf{supp.} \\
        \midrule
        % ============ Full Design ============
        \multirow{3}{*}{\textbf{Full}}
        & Panic      & -- & -- & 0.74 & 0.90 & 0.81 & 581  \\
        & No Panic   & -- & -- & 0.94 & 0.84 & 0.89 & 1130 \\
        & Avg.       & \textbf{0.86} & \textbf{0.87} & \textbf{0.84} & \textbf{0.87} & \textbf{0.85} & 1711 \\
        \cmidrule(lr){1-8}
        
        % ============ w/o RS, EA, MEA ============
        \multirow{3}{*}{\makecell[l]{w/o RS, EA, MEA}}
        & Panic      & -- & -- & 0.77 & 0.03 & 0.05 & 630  \\
        & No Panic   & -- & -- & 0.66 & 1   & 0.79 & 1183 \\
        & Avg.       & 0.66 & 0.51 & 0.72 & 0.51 & 0.42 & 1813 \\
        \cmidrule(lr){1-8}
        
        % ============ w/o EA, MEA ============
        \multirow{3}{*}{\makecell[l]{w/o EA, MEA}}
        & Panic      & -- & -- & 0.61 & 0.06 & 0.10 & 595  \\
        & No Panic   & -- & -- & 0.66 & 0.98 & 0.79 & 1105 \\
        & Avg.       & 0.66 & 0.52 & 0.63 & 0.52 & 0.44 & 1700 \\
        \cmidrule(lr){1-8}
        
        % ============ w/o MEA ============
        \multirow{3}{*}{w/o MEA}
        & Panic      & -- & -- & 0.56 & 0.76 & 0.64 & 605  \\
        & No Panic   & -- & -- & 0.85 & 0.69 & 0.76 & 1163 \\
        & Avg.       & 0.71 & 0.72 & 0.70 & 0.72 & 0.70 & 1768 \\
        \bottomrule
    \end{tabular}
    \caption{Ablation study, where the best performance is shown in \textbf{bold}. \textbf{RS} denotes \textit{Risk Sensing}, \textbf{EA} denotes \textit{Emotion Arousal}, and \textbf{MEA} denotes \textit{Multi-Expert Assessment}.}
    \label{tab:ablation}
\end{table}

\textbf{The synergy of RS, EA, and MEA:} Removing all three stages (w/o RS,EA,MEA) leads to severe degradation (accuracy: 0.66, AUC: 0.51), indicating their synergistic effects are critical, which severely impairs the model’s ability to capture users’ psychological traits and behavioral patterns.

\textbf{The foundational role of RS:} Removing EA and MEA (w/o EA,MEA) yields similar declines (accuracy: 0.66, AUC: 0.52), indicating RS’s foundational role in initializing context-aware risk perception to anchor psychological state recognition, also provides critical input for subsequent stages.

\textbf{The optimization role of MEA:} When only MEA component is removed (w/o MEA), the smaller declines (accuracy: 0.71, AUC: 0.72) reveal MEA’s optimization role in refining predictions, although its absence alone has a smaller impact compared to the joint absence of RS and EA.

Overall, these results validate the effectiveness of the PsychoAgent design. RS, EA and MEA work exhibit synergistic interactions, enabling comprehensive psychological modeling.

\subsection{Scalability Studies}

In this section, we investigate the interplay between model scale and performance under resource constraints by benchmarking four model sizes (14B, 32B, 72B, 671B) with/without the PsychoAgent framework. The experimental results are shown in Table ~\ref{tab:performance_comparison}, and our key findings are as follows.

\begin{itemize}[leftmargin=*]
    \vspace{-7pt}
    \item Pure parameter scaling fails to address data imbalance. All direct-prediction models achieve near-zero F1 on the Panic class (max 0.01 F1 for 671B), with macro-F1 stagnating at 0.39 from 14B to 72B, indicating inherent distributional bias toward majority classes;
    \vspace{-7pt}
    \item PsychoAgent empowers small models to surpass native large counterparts. The 14B model achieves 0.76 accuracy (+$\mathbf{18\%}$) and 0.43 Panic F1 (from zero), outperforming non-framework 32B/72B/671B models, demonstrating smaller models can transcend parameter limitations via task-driven reasoning under this framework;
    \vspace{-7pt}
    \item Framework efficacy exhibits scale-dependent synergy. The 671B model achieves peak accuracy (0.86, +$\mathbf{21\%}$), while the framework-enhanced 72B model (0.81 accuracy achieves $\mathbf{124.6\%}$ efficiency over the 671B baseline (0.65) at lower cost, proving mid-sized models offer cost-efficient alternatives. Notably, the 32B model underperforms 14B in Panic F1 (0.25 vs. 0.43), revealing framework benefits emerge only when the model size exceeds a certain threshold.
\end{itemize}

\begin{table}[ht]
    \centering
    \scriptsize % 将字体设置为更小尺寸
    \renewcommand{\arraystretch}{0.5} % 减小行间距
    \newcolumntype{C}{>{\centering\arraybackslash}p{0.8cm}} % 自定义居中列类型
    \newcolumntype{R}{>{\raggedleft\arraybackslash}p{0.9cm}} % 自定义右对齐列类型
    \setlength{\tabcolsep}{3pt} % 加宽列间距
    \begin{tabular}{@{} l l l l c c c c c @{}}
        \toprule
        \textbf{Model} & \textbf{Scale} & \textbf{Method} & \textbf{Class} & \textbf{Acc.} & \textbf{AUC} & \textbf{Prec.} & \textbf{Recall} & \textbf{F1} \\
        \midrule
        \multirow{6}{*}{\makecell[l]{Qwen 2.5-\\14B-Instruct}} & \multirow{6}{*}{14B} & \multirow{3}{*}{\makecell[l]{Direct\\Pred.}} & Panic & - & - & 0.00 & 0.00 & 0.00 \\
        &       &       & No Panic & - & - & 0.65 & 0.98 & 0.78 \\
        &       &       & Avg. & 0.64 & 0.49 & 0.32 & 0.49 & 0.39 \\
        &       & \multirow{3}{*}{\makecell[l]{psycho-\\Agent}} & Panic & - & - & 0.95 & 0.28 & 0.43 \\
        &       &       & No Panic & - & - & 0.74 & 0.99 & 0.84 \\
        &       &       & Avg. & 0.76 & \underline{0.64} & \textbf{0.84} & 0.64 & 0.64 \\
        \midrule
        \multirow{6}{*}{\makecell[l]{Qwen 2.5-\\32B-Instruct}} & \multirow{6}{*}{32B} & \multirow{3}{*}{\makecell[l]{Direct\\Pred.}} & Panic & - & - & 0.00 & 0.00 & 0.00 \\
        &       &       & No Panic & - & - & 0.65 & 1.00 & 0.79 \\
        &       &       & Avg. & 0.65 & 0.50 & 0.33 & 0.50 & 0.39 \\
        &       & \multirow{3}{*}{\makecell[l]{psycho-\\Agent}} & Panic & - & - & 1.00 & 0.14 & 0.25 \\
        &       &       & No Panic & - & - & 0.68 & 1.00 & 0.81 \\
        &       &       & Avg. & 0.70 & 0.57 & \textbf{0.84} & 0.57 & 0.53 \\
        \midrule
        \multirow{6}{*}{\makecell[l]{Qwen 2.5-\\72B-Instruct}} & \multirow{6}{*}{72B} & \multirow{3}{*}{\makecell[l]{Direct\\Pred.}} & Panic & - & - & 0.00 & 0.00 & 0.00 \\
        &       &       & No Panic & - & - & 0.65 & 1.00 & 0.79 \\
        &       &       & Avg. & 0.65 & 0.50 & 0.33 & 0.50 & 0.39 \\
        &       & \multirow{3}{*}{\makecell[l]{psycho-\\Agent}} & Panic & - & - & 0.86 & 0.31 & 0.45 \\
        &       &       & No Panic & - & - & 0.81 & 0.98 & 0.89 \\
        &       &       & Avg. & \underline{0.81} & \underline{0.64} & \underline{0.83} & \underline{0.65} & \underline{0.67} \\
        \midrule
        \multirow{6}{*}{\makecell[l]{DeepSeek\\-V3}} & \multirow{6}{*}{671B} & \multirow{3}{*}{\makecell[l]{Direct\\Pred.}} & Panic & - & - & 0.50 & 0.00 & 0.01 \\
        &       &       & No Panic & - & - & 0.65 & 1.00 & 0.79 \\
        &       &       & Avg. & 0.65 & 0.50 & 0.60 & \underline{0.65} & 0.52 \\
        &       & \multirow{3}{*}{\makecell[l]{psycho-\\Agent}} & Panic & - & - & 0.74 & 0.90 & 0.81 \\
        &       &       & No Panic & - & - & 0.94 & 0.84 & 0.89 \\
        &       &       & Avg. & \textbf{0.86} & \textbf{0.87} & \textbf{0.84} & \textbf{0.87} & \textbf{0.85} \\
        \bottomrule
    \end{tabular}
    \caption{Performance Comparison of Different Model Sizes, where the best performance is shown in \textbf{bold} and the second best is \underline{underlined}.}
    \label{tab:performance_comparison}%
\end{table}%

Overall, the results indicate that model size and performance lack a simple linear relationship. PsychoAgent enables compact models to rival or exceed larger counterparts in resource-constrained scenarios through psychological modeling. 

Additionally, to visually demonstrate our framework's superior mechanistic interpretability compared to traditional data-fitting approaches, we present two representative prediction cases. Detailed results are in Appendix ~\ref{subsec:case_studies}.

\section{RELATED WORK}
We discuss the related works from three aspects: panic arousal theory, current panic analysis methods, and LLM-based role-playing approaches.

\noindent\textbf{Panic Emotion Arousal Theory.}
Psychological studies establish panic as an uncertainty-driven fear linked to risk perception \cite{weigang2024evolution}, shaped by event-individual trait interactions \cite{davis2020transformation,syrdal2018engagement}. Traditional methods (e.g., surveys) suffer from subjective bias and sampling limitations, failing to track real-time risk cognition in social media contexts. Our work bridges this gap by integrating psychological arousal theory with multi-domain data modeling, advancing computational panic analysis through mechanism-driven frameworks.

\noindent\textbf{Panic Emotion Analysis.}
Current panic emotion research relies on event-driven static labeling and keyword filtering \cite{mitrovic2024comparing,weigang2024evolution}, focusing on post-disaster data while lacking baseline behavioral data for prediction. Existing models, constrained to single-domain text analysis \cite{mitrovic2024comparing,HariharanDWA17}, exhibit limited interpretability due to insufficient multi-domain feature integration and inability to deconstruct panic formation pathways. We address these gaps via a dual-phase dataset and psychology-anchored LLM agent framework for interpretable panic pathway modeling.

\noindent\textbf{LLM-Based Personalized Role-Playing.}
Personalized role simulation typically employs parameterized fine-tuning or prompt engineering. Prior work employs data-intensive training for fictional/historical role replication \cite{li2023chatharuhi,wang2023rolellm,ran2024capturing}, yet struggles with social media's data sparsity and privacy constraints \cite{abbasiantaeb2024let}. Non-parametric prompting methods \cite{park2024generative,jiang2023evaluating} reduce data dependency but produce semantically shallow responses lacking psychological depth. We propose a CoT-driven LLM approach to simulate panic response logic via risk-emotion causal chains, enabling lightweight, mechanism-driven role modeling for social media.

%\vspace{-10pt}
\section{Conclusions}

We introduce PsychoAgent, a novel framework for predicting dynamic panic emotion in social media users during disasters, grounded in psychological emotion arousal theory. PsychoAgent uniquely integrates a human-LLM collaboratively annotated dataset for fine-grained emotion analysis, a psychology-driven feature fusion mechanism to model public panic dynamics, and LLM-based agent simulations to trace panic formation pathways, moving beyond traditional reliance on explicit features. Experiments demonstrate that PsychoAgent significantly outperforms existing baselines in both prediction accuracy and interpretability. Our findings have important implications for developing explainable AI systems that offer deeper insights into collective emotional dynamics during crises. Future work could expand this framework to investigate panic propagation dynamics across social networks and to develop real-time intervention strategies to mitigate emotional contagion during emergencies.

% \clearpage
\section{Limitations} 
This work primarily focus on individual-level panic prediction by simulating psychological mechanisms. While this approach effectively models cognitive-emotional chains and yields explainable predictions, it is not without limitations.
First, LLM hallucinations may cause deviations from psychological priors in modeling risk perception and emotion arousal. Future work should integrate more robust self-correction mechanisms within the agents to mitigate these discrepancies. Second, stylistic differences between LLM-generated panic texts and authentic user posts can create semantic divergence, potentially leading to false negatives in our self-trained BERT classifier. Enhancing the linguistic realism of generated text to better align with human expression is an important next step. Furthermore, mainstream LLMs (e.g., Deepseek, GPT) often impose political correctness guardrails that suppress negative emotional expressions, potentially underdetecting panic users, suggesting our framework is better suited for models without such constraints.

\section{Ethics Statement}
% In the data collection, we ensure there is no identifiable information about individuals (faces, license plates) or private properties.
The datasets used in this work are all publicly available, so there is no ethical concern.

% \input{doc/7_acknow}
% \clearpage

\bibliography{custom}

% \documentclass{book}
% \usepackage[utf8]{inputenc}
% \usepackage{hyperref}
% \usepackage{graphicx}
% \usepackage{booktabs}
% \usepackage{multirow}

% \begin{document}

\appendix

\clearpage
\section{Appendix}
\label{sec:Appendix}
\subsection{Comparison of Current Emotion Prediction Works}
\label{subsec:current_emotion_prediction}
We summarize the research landscape of existing emotion prediction approaches and draw a comparison with our proposed framework by examining aspects including dataset construction, application scenarios, theoretical foundations, multi-domain feature fusion, generative model architectures, interpretability, and granularity of emotion prediction. The comparative results are presented in Table ~\ref{tab:comparison}.

\subsection{Dataset Labeling and Self-trained BERT Model}
\label{subsec:Dataset_BERT}

Focusing on panic prediction during sudden disasters, our dataset centers on post-disaster texts with panic emotion annotations. We develop a hybrid annotation framework combining LLM-assisted labeling (1,065 entries), human crowdsourcing, EDA-based data augmentation, and BERT fine-tuning for automated tagging. Statistical details of the dataset is presented in Table \ref{tab:dataset_detail}. Specific implementation details are provided in subsequent sections.

\begin{table}[htbp]
\scriptsize % 将字体设置为更小尺寸
\setlength{\extrarowheight}{8pt} % 增加额外行高
\renewcommand{\arraystretch}{0.85} % 适当增加行间拉伸系数
\setlength{\extrarowheight}{4pt} % 增加额外行高
  \centering
  \begin{tabularx}{0.95\columnwidth}{@{}>{\centering\arraybackslash}X >{\centering\arraybackslash}X@{}}
    \toprule
    \textbf{Category} & \textbf{Quantity} \\
    \midrule
    Panic User Count & 3,240 \\
    Non-Panic User Count & 5,825 \\
    \textbf{Total Users} & \textbf{9,065} \\
    \addlinespace[1pt]
    Pre-Disaster Tweet Count & 242,363 \\
    Post-Disaster Tweet Count & 1,142,626 \\
    \textbf{Total Tweets} & \textbf{1,384,989} \\
    \bottomrule
  \end{tabularx}
  \caption{Details of the constructed dataset.}
  \label{tab:dataset_detail}
\end{table}

\subsubsection{Data preprocessing procedures}
\label{subsubsec:Data_preprocessing}
Our human-LLM collaboratively annotated Hurricane Sandy dataset innovatively integrates panic emotion labels across pre-disaster and post-disaster phases. Before annotation, we preprocessed the collected dataset as follows:

\begin{enumerate}
[leftmargin=*,itemsep=2pt,topsep=0pt,parsep=0pt]
 \item Data Filtering: Removed malformed, highly redundant (cosine similarity $>$0.85), or invalid entries (posts with $<$5 meaningful tokens);
 \item Irrelevant Content Removal: Leveraged human-LLM collaboration to exclude non-Sandy-related texts from the keyword-crawled public dataset(similar to Section 3.2);
 \item User Selection: Retained users who posted in both phases, with $\geq$10 pre-disaster tweets;
 \item Text Sanitization: Stripped URLs, retweets ("RT @xxx:"), and non-alphanumeric characters;
 \item Temporal Stratification: Split the dataset into pre-disaster and peri-disaster subsets based on timestamps.
\end{enumerate}

\subsubsection{LLM-Assisted Annotation Prompts}
\label{subsubsec:LLM_Annotation_Prompt}
The ChatGPT-4o assisted annotation prompts used in our dataset labeling process are shown as follows. Since LLMs are trained on extensive corpus data, they inherently contain general knowledge about Hurricane Sandy and panic emotions. Our approach simply required prompting the LLM to leverage its existing knowledge to determine whether input texts: (1) relate to Hurricane Sandy, and (2) exhibit user panic emotions.

%\begin{figure}[h]
  %\centering
  %\includegraphics[width=\linewidth]{figure/5.png}
  %\caption{Path of Panic Formation}
  %\label{fig:annotation_prompt}
%\end{figure}

%\vspace{0.5em}
\begin{figure}
  \begin{tcolorbox}[%
  title={\textbf{Prompt for LLM-assisted annotation\\(Hurricane Related)}},  % 标题加粗
  label={box:hurricane_related},
  fontupper=\scriptsize,        % 内容区字号
  colback=blue!5,         % 背景色
  colframe=blue!75!black,  % 边框颜色
  fonttitle=\bfseries\large,     % 标题字体加粗大号字
]
\textbf{Question prompt:}
You are a helpful assistant that identifies text relevance to Hurricane Sandy using specific keywords and related terms. Evaluate the following text for relevance to Hurricane Sandy. Focus only on the text provided and check for keywords such as "Sandy", "fierce winds", "heavy rainfall", "flooding", "long power outages", "population centers", "target area", "economic loss", "hurricane", "storm", "superstorm", "tropical cyclone", and other disaster-related terms. If you find any relevant keywords, respond with "Yes" and provide an explanation. If not, respond with "No" and explain why it's not relevant.
Input text is: Creepy clouds. Bad weather here in Chi...See the face?? Go back!!!!\\
\hspace*{\fill}\\
\textbf{LLM answer:}
Yes, the text is relevant to Hurricane Sandy. It contains the keyword "Hurricane" and specifically mentions "Sandy", which directly relates to Hurricane Sandy itself. Additionally, it provides information about the hurricane's location (26.3N 76.9W) and its movement (NW at 11 mph), which are pertinent details regarding the storm's status and trajectory. This aligns well with the context of discussing Hurricane Sandy and its impact.
\end{tcolorbox}
    \centering
    %\includegraphics[width=0.5\linewidth]{}
    %\caption{\textbf{Prompt for LLM-assisted annotation\\(Panic Related)}}
    %\label{box:hurricane_related}
\end{figure}

%\vspace{0.5em}

\begin{figure}
  \begin{tcolorbox}[%
  title={\textbf{Prompt for LLM-assisted annotation\\(Panic Related)}},  % 标题加粗
  label={box:panic_related},
  fontupper=\scriptsize,        % 内容区字号
  colback=blue!5,         % 背景色
  colframe=blue!75!black,  % 边框颜色
  fonttitle=\bfseries\large    % 标题字体加粗
%  width=\textwidth       % 设置宽度为当前栏宽

]
\textbf{Question prompt:}
You are a helpful assistant that evaluates text for emotional content, particularly focusing on panic emotions. You will analyze the input text based on the context and linguistic cues to determine if it reflects panic emotions from the text's author. If panic emotions are present, respond with "Yes" and provide an explanation. If panic emotions are not present, respond with "No" and explain why.
Input text is: Creepy clouds. Bad weather here in Chi...See the face?? Go back!!!!\\
\hspace*{\fill}\\
\textbf{LLM answer:}
Yes. The text reflects panic emotions through several linguistic cues. Phrases like "Creepy clouds" and "Bad weather" indicate a sense of unease and fear about the current situation. The exclamation "See the face??" suggests a heightened emotional response, ponssibly idicating a perceived threat or danger. The command "Go back!!!!" further emphasizes urgency and panic, as it implies a strong desire to escape or avoid a potentially harmful situation. Overall, the combination of unsettling descriptions and urgent commands conveys a clear sense of panic from the author.
\end{tcolorbox}
    \centering
    %\includegraphics[width=0.5\linewidth]{}
    %\caption{\textbf{Prompt for LLM-assisted annotation\\(Panic Related)}}
    %\label{box:panic_related_box}
\end{figure}

%\definecolor{presence-color}{RGB}{239, 247, 234}
%\definecolor{absence-color}{RGB}{255 228 225}
\renewcommand{\tabularxcolumn}[1]{m{#1}} % 将X列的垂直对齐改为居中

\begin{table*}[htbp]
\scriptsize % 将字体设置为更小尺寸
\centering
\renewcommand{\arraystretch}{0.85} % 增加行高保证垂直居中
\resizebox{\textwidth}{!}{%
\begin{tabularx}{\linewidth}{ 
  >{\hsize=1.1\hsize\centering\arraybackslash}X % Ref.
  >{\hsize=1.0\hsize\centering\arraybackslash}X  % Dedicated Dataset
  >{\hsize=0.9\hsize\centering\arraybackslash}X % Scenario 
  >{\hsize=0.9\hsize\centering\arraybackslash}X  % Psychology
  >{\hsize=0.8\hsize\centering\arraybackslash}X % Features
  >{\hsize=1.5\hsize\centering\arraybackslash}X % Architecture
  >{\hsize=1.0\hsize\centering\arraybackslash}X  % Interpretability
  >{\hsize=0.7\hsize\centering\arraybackslash}X  % Granularity
}  
\toprule
\textbf{Ref.} & \textbf{Dataset} & \textbf{Scenario} & \textbf{Psychology} & \textbf{Features} & \textbf{Generative Architecture} & \textbf{Interpretability} & \textbf{Granularity} \\
\midrule
\cite{yongsatianchot2023s} & / & Sports competition & / & Contextual & ChatGPT-4 & Partial (Output) & -- \\
\cite{regan2024can} & EmotionBench & Multi-scenario stories & Emotion assessment & Contextual & GPT-3.5-Turbo & Partial (Context) & Binary \\
\cite{mou2024multi} & / & News & / & Cross-doc & / & / & Ekman-6 \\
\cite{li2020interactive} & / & Text dialogues & / & / & / & / & Ekman-6 /4-class \\
\cite{gao2024enhancing} & / & News & / & / & RoBERTa+ChatGPT & Partial (Explanation-Assisted) & 8-class \\
\cite{houlihan2023emotion} & Game data & "Split or Steal" & Intuitive psychology & Preferences & / & Full-chain & 20-class \\
\cite{alsaedi2022transfer} & / & Social media & / & / & / & / & 5-class \\
Ours & Disaster & Emergencies & Risk perception + Emotional arousal & Multi-domain fusion & LLM+MoE & Full-chain & Panic-specific \\
\bottomrule
\end{tabularx}  
}
\caption{Comparative summary of emotion prediction research status. Dedicated dataset (whether creating task-specific data), Scenario (application in specific contexts), Psychology (integration of psychological theories), Features (multi-domain feature incorporation), Generative architecture (use of generative models), Interpretability (model transparency), and Granularity (emotion classification specificity). \textit{Cells marked with a slash (/) indicate the absence of specified design components.}}
\label{tab:comparison} 
\end{table*}

%\vspace{0.5em}
\subsubsection{Human Crowdsourcing Verification Rules} 
\label{subsubsec:Crowdsourcing_rules}

\begin{figure}[htbp]
  \centering
  \includegraphics[width=\linewidth]{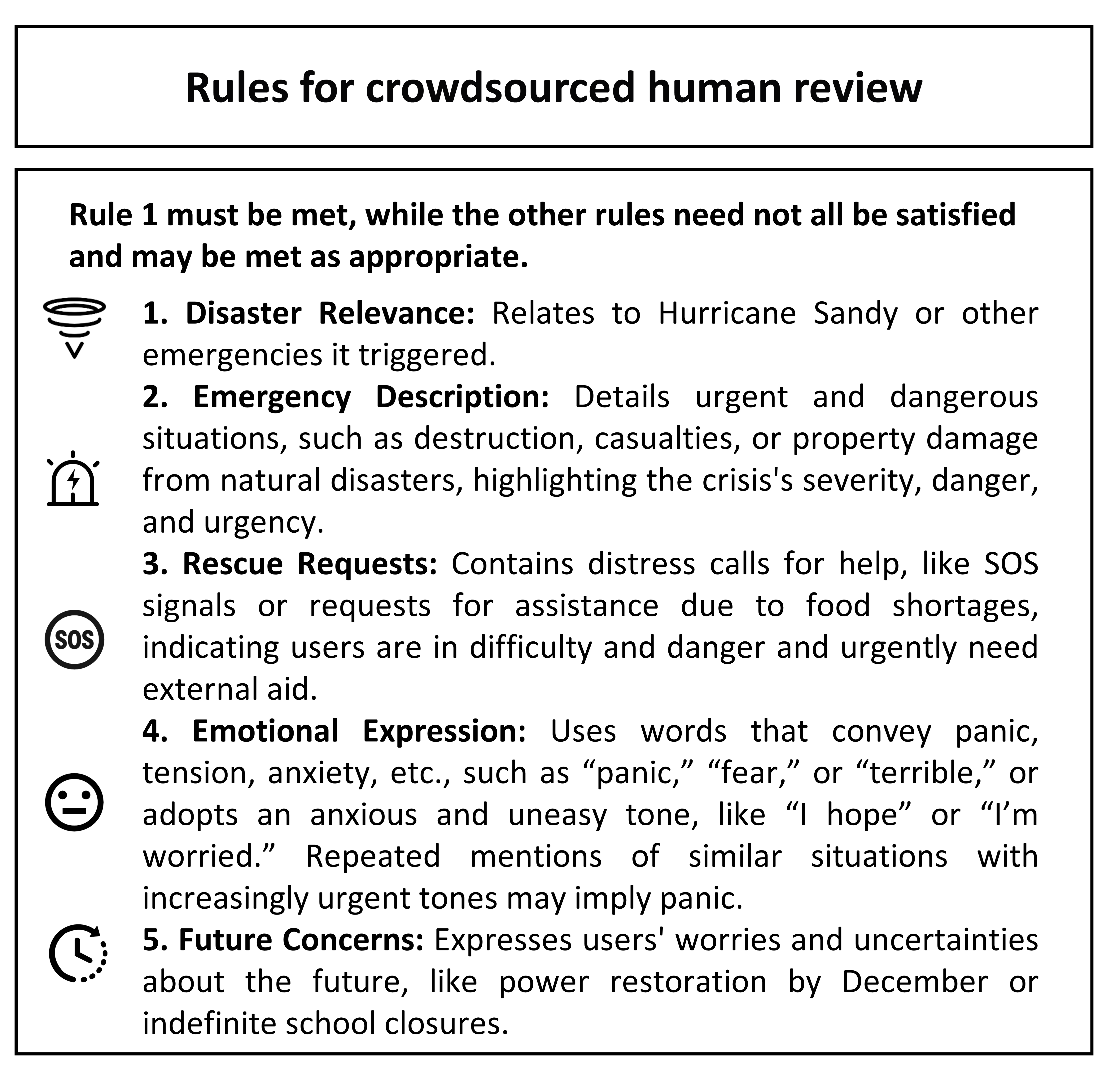}
  \caption{Crowdsourced validation protocol for panic emotion annotation.}
  \label{fig:validation_rules}
\end{figure}

Based on psychological studies of panic emotions and the specific context of Hurricane Sandy, we establish the crowdsourced validation rules for panic emotion annotation as shown in Figure~\ref{fig:validation_rules}. The rules operationalize panic emotion manifestations through five key dimensions.

\subsubsection{BERT Model Training Effect}
\label{subsubsec:BERT_training}
We fine-tuned a BERT model on an augmented and balanced dataset for auto-labeling remaining data, which also provide a reliable benchmark tool for subsequent emotion discrimination in generated texts. To validate training efficacy, we tested the model on both the augmented dataset and the original human-verified dataset, with results (Table ~\ref{tab:bert_augmented} and Table ~\ref{tab:bert_imbalanced}{}) demonstrating robust performance even on non-augmented data, confirming the effectiveness of our EDA enhanced training strategy.

\begin{table}[ht]
  \centering
  \scriptsize % 将字体设置为更小尺寸
    \renewcommand{\arraystretch}{0.4} % 减小行间距
    \setlength{\tabcolsep}{9pt} % 压缩列间距
    %\newcolumntype{C}{>{\centering\arraybackslash}p{0.8cm}} % 自定义居中列类型
    %\newcolumntype{R}{>{\raggedleft\arraybackslash}p{0.9cm}} % 自定义右对齐列类型
    \begin{tabular}{lccccc}
    \toprule
          & Precision & Recall & F1-score & Support \\
    \midrule
    No Panic & 1.00     & 0.99   & 0.99     & 513 \\
    Panic     & 0.99     & 1.00   & 0.99     & 497 \\
    \midrule
    Accuracy  &          &        & 0.99 & 1010 \\
    \midrule
    Macro Avg & 0.99     & 0.99   & 0.99     & 1010 \\
    Weighted Avg & 0.99  & 0.99   & 0.99     & 1010 \\
    \bottomrule
    \end{tabular}%
  \caption{Training Effect of BERT Model on Augmented Data.}
  \label{tab:bert_augmented}%
\end{table}

\begin{table}[ht]
  \centering
  \scriptsize % 将字体设置为更小尺寸
  \renewcommand{\arraystretch}{0.4} % 减小行间距
  \setlength{\tabcolsep}{9pt} % 压缩列间距
    \begin{tabular}{lccccc}
    \toprule
          & Precision & Recall & F1-score & Support \\
    \midrule
    No Panic & 1.00     & 0.99   & 1.00     & 1002 \\
    Panic     & 0.90     & 1.00   & 0.95     & 63 \\
    \midrule
    Accuracy  &          &        & 0.99 & 1065 \\
    \midrule
    Macro Avg & 0.95     & 1.00   & 0.97     & 1065 \\
    Weighted Avg & 0.99  & 0.99   & 0.99     & 1065 \\
    \bottomrule
    \end{tabular}%
  \caption{Testing Effect of BERT Model on Original Imbalanced Data.}
  \label{tab:bert_imbalanced}%
\end{table}

\subsection{Details of Individual Feature Extraction}
\label{subsec:individual_freature}
\subsubsection{Extraction of the Big Five Personality Traits} 
\label{subsubsec:big_five_traits}
The bert-base-personality model, fine-tuned on a curated personality dataset from the BERT-BASE-UNCASED backbone, predicts Big Five personality traits (Openness, Extraversion, Neuroticism, Agreeableness, Conscientiousness) via transfer learning, outputting a dictionary of normalized scores (0–1) for each dimension. While the model lacks disclosed accuracy/F1 metrics, we validated its consistency by testing it twice on $50\%$ splits of users’ pre-disaster texts, observing $>95\%$ intra-user trait alignment within acceptable variance thresholds, confirming its reliability for personality profiling. The personality consistency verification results are shown in the Table ~\ref{tab:consistency}.

% \usepackage{multirow}
\begin{table}[]
\scriptsize % 设置小号字体
\renewcommand{\arraystretch}{1.5} % 减小行间距
\setlength{\tabcolsep}{13pt} % 压缩列间距
\begin{tabular}{cccc}
\hline
\multirow{2}{*}{\textbf{\begin{tabular}[c]{@{}c@{}}Floating\\ range\end{tabular}}} & \multicolumn{2}{c}{\textbf{User Count/Proportion}} & \multirow{2}{*}{\textbf{\begin{tabular}[c]{@{}c@{}}Total \\ Users\end{tabular}}} \\ \cline{2-3}
    & Consistent  & Inconsistent   &  \\ \hline
15\%  & 8880 (97.96\%) & 185 (2.04\%)  & 9065 \\
20\%  & 9032 (99.64\%) & 33 (0.36\%)   & 9065  \\ \hline
\end{tabular}
\caption{Consistency Analysis of Bert-base-personality Personality Detection Model.}
\label{tab:consistency}
\end{table}

\subsubsection{Focused Topic Feature Extraction}
\label{subsubsec:topic_feature}

The LDA model, a generative Bayesian probabilistic algorithm with a three-layer structure (word, topic, corpus), includes two key parameters: the number of topics and keywords. 

In this study, we configure the LDA model to cluster all users’ pre-disaster posts into 25 distinct topics, with each topic characterized by 10 keywords. This process also identifies the most likely topic category for each post. This process is formalized as follows:
\begin{equation}
\theta^{k},\ \phi_{k} = \psi_{\text{topic}}(I^{i}_{t},\ k=25,\ n=10),
\end{equation}
where $\theta^{k} \in \mathbb{R}^{25}$ is the 25 topics generated by LDA; $\phi_k$ is the keyword list for the $k$-th topic; $\psi_{\text{topic}}$ is the topic clustering model.

Subsequently, we prompt ChatGPT-4o to merge similar topics based on their summarized keywords, ultimately consolidating them into 8 common thematic categories (see Table ~\ref{tab:pre_disaster_topics} for more details). This process is formalized as:
\begin{equation}
\Gamma=LLM_{topic}(\{\phi_k\}_{k=1}^{25},p_\Gamma),
\end{equation}
where $p_{\mathrm{\Gamma}}$ is the prompt used to instruct ChatGPT-4o for topic induction; $LLM_{topic}$ is the ChatGPT-4o-based topic consolidation model, which outputs 8 consolidated themes; $\Gamma$ is the membership relationships from the original 25 topics to the 8 consolidated themes.

\begin{table}[ht]
    \centering
    \scriptsize 
    \renewcommand{\arraystretch}{0.6} % 行间距
    \setlength{\tabcolsep}{2.5pt}       % 列间距
    \begin{tabularx}{\linewidth}{@{}>{\raggedright}p{3cm} X@{}} % 列宽配置
        \toprule
        \textbf{Topic Category} & \textbf{Top Keywords} \\
        \midrule
        Politics \& Elections & debate, obama, presidential, governor, business \\
        Natural Disasters \& Weather & weather, wind, rain, hurricane, storm \\
        Energy \& Environment & power, solar, gas, energy, climate \\
        Sports \& Entertainment & york, giants, jets, nfl, yankees \\
        Economy \& Business & gas, prices, obama, economy, wall \\
        Society \& News & news, governor, business, china, friends \\
        Technology \& Innovation & power, solar, tech, play, technology \\
        Miscellaneous & hurricane, tropical, newyork, storm, east \\
        \bottomrule
    \end{tabularx}
    \caption{Social Media Users' Event Topics of Interest Before Disasters}
    \label{tab:pre_disaster_topics}
\end{table}

Then, we aggregate each user's pre-disaster topical focuses based on the topic labels assigned to their historical posts prior to the disaster, which can be expressed as:
\begin{equation}
\tau^i=\Gamma\cdot\Theta^i,
\end{equation}
where $\Theta^{i}$ is the topic vector of the $i$-th user on the LDA-clustered topics; $\Gamma$ is the membership relationships from the LDA-clustered topics to the LLM-consolidated themes; $\tau^{i}$ is the summarized thematic focus vector of the same user on the 8 consolidated themes, represented as a list of theme names.

\subsubsection{Tone Feature Extraction}
\label{subsubsec:Tone_feature}

Leveraging the capabilities of LLM in natural language processing (NLP) and text pattern recognition, we design a specialized prompt that extracts and condenses each user’s linguistic tone features. Specifically, we instruct ChatGPT-4o to analyze the linguistic tone of each user’s posts under normal conditions, returning three words to describe their tone features, which ensuring efficient and interpretable stylistic profiling. The prompt template for tone feature induction is detailed as follows.

%\begin{figure}[htbp]
%  \centering
  %\includegraphics[width=\linewidth]{figure/7.png}
  %\caption{Prompt and Example for Tone Feature Extraction}
  %\label{fig:Tone_feature_prompt}
%\end{figure}

\begin{figure}

  \begin{tcolorbox}[%
  title={\textbf{Prompt for Tone Feature Extraction}},  % 标题加粗
  label={box:Tone_feature},
  fontupper=\scriptsize,        % 内容区字号
  colback=blue!5,         % 背景色
  colframe=blue!75!black,  % 边框颜色
  fonttitle=\bfseries\large,     % 标题字体加粗
]
\textbf{Question prompt:}
You are a professional linguist expert who can identify the tone of voice in social media texts. A social media user has sent the following tweets: {row\lbrack{"text"}\rbrack}. Describe this user’s overall tone of voice on the social media with three words. Only output in the exact format: xxx, xxx, xxx."\\
\hspace*{\fill}\\
\textbf{LLM answer:}
Casual, Humorous, Restless

\end{tcolorbox}
    \centering
    %\includegraphics[width=0.5\linewidth]{}
    %\caption{\textbf{Prompt for LLM-assisted annotation\\(Panic Related)}}
    %\label{box:Tone_feature}
\end{figure}

\begin{table*}[ht]
  \centering
  \scriptsize % 调整字体大小为较小尺寸
  \renewcommand{\arraystretch}{0.6} % 调整行间距，使其更紧凑
  \setlength{\tabcolsep}{3pt} % 调整列间距
  \begin{tabular}{p{5cm}p{11cm}} % 加宽列宽
    \toprule
    \textbf{Knowledge Category} & \textbf{Content} \\
    \midrule
    
    Public Risk Perception Formation & 
    \begin{itemize}[leftmargin=*,nosep]
        \item Risk perception is shaped by two factors and their interaction: 
        \begin{itemize}[leftmargin=*,nosep]
            \item[a)] Characteristics of the risk event itself
            \item[b)] Personal characteristics of the audience
        \end{itemize}
    \end{itemize} \\
    %\addlinespace[0.5em]
    
    \cmidrule(r){1-2}
    Personality Traits and Risk Response & 
    \begin{itemize}[leftmargin=*,nosep]
        \item High Psychoticism: Associated with overestimation of event controllability
        \item High Extraversion: Correlates with perceived understanding of emergencies (e.g., pandemic knowledge)
        \item High Neuroticism: Linked to lower emergency comprehension and higher fear levels
        \item Extraverts tend to adopt proactive measures
        \item Emotionally unstable individuals (high Neuroticism) prefer passive coping strategies
    \end{itemize} \\
    %\addlinespace[0.5em]
    
    \cmidrule(r){1-2}
    Social Media Language Style Effects & 
    \begin{itemize}[leftmargin=*,nosep]
        \item Sarcasm/irony may amplify anxiety in crisis contexts
    \end{itemize} \\
    %\addlinespace[0.5em]
    
    \cmidrule(r){1-2}
    Content Type Emotional Impacts & 
    \begin{itemize}[leftmargin=*,nosep]
        \item Disaster-related serious news increases situational awareness but may elevate stress
    \end{itemize} \\
    %\addlinespace[0.5em]
    
    \cmidrule(r){1-2}
    Emotional Stability Mechanisms & 
    \begin{itemize}[leftmargin=*,nosep]
        \item Regular use of cognitive reappraisal strategies buffers acute stress during disasters
    \end{itemize} \\
    %\addlinespace[0.5em]
    
    \cmidrule(r){1-2}
    Social Media Network Property Roles & 
    \begin{itemize}[leftmargin=*,nosep]
        \item Users with more follows/followers are more likely to be exposed to diverse and potentially conflicting information, which can increase cognitive load and anxiety
        \item Dense social networks (many friends) can lead to group polarization and echo chamber effects, amplifying panic through frequent interactions
        \item Social comparison on platforms with many users can weaken self-efficacy when others display superior coping resources
    \end{itemize} \\
    \bottomrule
  \end{tabular}
\caption{Psychological Knowledge Embedded Content.}
\label{tab:psych_knowledge}
\end{table*}

\subsection{CoT-Driven LLM-Based Agent}
\label{subsec:CoT_driven_Agent}
\subsubsection{Psychological Knowledge Injection}
\label{subsubsec:psycho_konwledge}

\begin{enumerate}[label=(\arabic*),itemsep=2pt,topsep=0pt,leftmargin=*]
  \item Psychological knowledge comprehension: we integrated a psychology knowledge system with six core modules, which covers public risk perception formation, personality traits and risk response (with a focus on the Big Five model), social  media language style effects, content type emotional impacts, emotional stability mechanisms, social media network property roles, establishing the theoretical foundation and baseline cognitive judgment for psychology-driven prediction in subsequent phases (detailed in Table ~\ref{tab:psych_knowledge});
  \item Disaster event data analysis: Integrating real-time meteorological data to form a dynamic risk perception of disasters, analyzing the physical features and linking to the "risk event features" in psychology;
  \item User profile initialization: Using dual modeling for the agent's initialization. A basic profile combines user static traits (e.g., Big Five personality traits) with risk communication analysis (e.g., geographical location, network topology, and behavioral frequency). Additionally, the Contriever model retrieves the top 5 hurricane-related posts from users’ pre-disaster posts as supplementary material, helping the agent infer latent behavioral tendencies for a more accurate individual simulation.
\end{enumerate}

\begin{table*}[htbp]
\centering
\scriptsize % 调整字体大小为较小尺寸
  \renewcommand{\arraystretch}{0.6} % 调整行间距，使其更紧凑
\begin{tabularx}{\textwidth}{>{\raggedright\arraybackslash}p{2cm}X}
\toprule

\multicolumn{2}{l}{\textbf{Knowledge \& Awareness (KA) sub-scale}} \\
\midrule[0.8pt]
1 & I am familiar with the natural hazard/disaster preparedness materials relevant to my area. \\
2 & I know which household preparedness measures are needed to stay safe in a natural hazard/disaster. \\
3 & I know how to adequately prepare my home for the forthcoming fire/flood/cyclone season. \\
4 & I know what to look out for in my home and workplace if an emergency weather situation should develop. \\
5 & I am familiar with the disaster warning system messages used for extreme weather events. \\
6 & I am confident that I know what to do and what actions to take in a severe weather situation. \\
7 & I would be able to locate the natural hazard/disaster preparedness materials in a warning situation easily. \\
8 & I am knowledgeable about the impact that a natural hazard/disaster can have on my home. \\
9 & I know what the difference is between a disaster warning and a disaster watch situation. \\
10 & I am familiar with the weather signs of an approaching fire/flood/cyclone. \\

%\addlinespace[1em]
\midrule
\multicolumn{2}{l}{\textbf{KA construct reliability}} \\
\multicolumn{2}{l}{\textbf{Anticipation, Awareness \& Management (AAM) sub-scale}} \\
\midrule[0.8pt]
1 & I think I am able to manage my feelings pretty well in difficult and challenging situations. \\
2 & In a natural hazard/disaster situation I would be able to cope with my anxiety and fear. \\
3 & I seem to be able to stay cool and calm in most difficult situations. \\
4 & I feel reasonably confident in my own ability to deal with stressful situations that I might find myself in. \\
5 & When necessary, I can talk myself through challenging situations. \\
6 & If I found myself in a natural hazard/disaster situation I would know how to manage my own response to the situation. \\
7 & I know which strategies I could use to calm myself in a natural hazard/disaster situation. \\
8 & I have a good idea of how I would likely respond in an emergency situation. \\

\bottomrule
\end{tabularx}
\caption{Psychological Preparedness for Disaster Threats Scale (PPDTS).}
\label{tab:assessment_scales}
\end{table*}

\subsubsection{Psychological Preparedness for Disaster Threats Scale (PPDTS)}
\label{subsubsec:PPTDS}
The Perceived Preparedness for Disaster Threat Scale (PPDTS) exhibits a two-factor structure—Knowledge \& Awareness (KA) and Anticipation, Awareness \& Management (AAM)—with strong inter-component correlations and high internal consistency reliability for both the full scale and subscales. Its 18 scenario-based items align with specific theoretical constructs from psychological knowledge modules (see Table ~\ref{tab:assessment_scales}). For instance, the item "I am confident that I know what to do and what actions to take in a severe weather situation." activates the agent’s long-term memory of the interplay between sense of control and user extraversion, enabling dynamic psychological computation.

\subsubsection{Tweet generation and multi-expert evaluation}
\label{subsubsec:MoE_system}
To ensure generated tweets align with user personality traits while reflecting psychological dynamics in disaster events, we prompt the agent to perform the following reasoning steps in sequence, thus building a multi - source memory integration mechanism. Specifically:
 
\begin{enumerate}[label=(\arabic*),itemsep=2pt,topsep=0pt,leftmargin=*]
  \item Extract psychological domain knowledge as constraints;
  \item Inject key situational parameters from real-time hurricane data;
  \item social network behavioral traits and personal traits from user profiles; 
  \item Integrate risk perception features derived from Phase 2 assessments;
  \item Determine emotional tone via panic arousal probability values;
  \item Generate the top 3 most probable posts the user would publish.
\end{enumerate}

To assess the generated text's consistency and validity across psychological compatibility, linguistic coherence, factual reliability, and emotional rationality, we use prompt engineering to coordinate joint evaluation by four domain-specific expert models:

\begin{enumerate}[label=(\arabic*),itemsep=2pt,topsep=0pt,leftmargin=*]
  \item The psychology expert model: Verifies alignment between generated texts and user psychological profiles (particularly Big Five personality traits);
  \item The linguistic expert model: Analyzes linguistic style coherence with historical posts;
  \item The factual consistency model: Ensures disaster-event relevance and accuracy; 
  \item The emotional alignment model: Maintains dynamic alignment between textual emotional intensity (via lexical analysis) and user panic probability.
\end{enumerate}

\subsubsection{Prompts and Case Demonstrations in Different Phases}
\label{subsubsec:case_demonstrations}
Our proposed prompting methods are as shown below. Using a user near Florida as an example, we employ CoT to build an agent with long-term memory. This agent simulates the same disaster scenario for the target user, aiming to predict potential panic emotion immediately before or during a hurricane's landfall.

%\begin{figure}[htbp]
  %\centering
  %\includegraphics[width=\linewidth]{figure/8.png}
  %\caption{Demonstration of Multi-Stage Prompting in CoT-Driven Agent for Simulating User Panic Emotion Prediction}
  %\label{fig:case_demostration}
%\end{figure}

\begin{figure*}
  \begin{tcolorbox}[%
  title={\textbf{Stage 1: Disaster Perception Stage}},  % 标题加粗
  label={box:stage_1},
  fontupper=\scriptsize,        % 内容区字号
  colback=blue!5,         % 背景色
  colframe=blue!75!black,  % 边框颜色
  fonttitle=\bfseries\large,     % 标题字体加粗
]
\textbf{Question prompt:}
You are a psychologist specializing in predicting public emotional trends during emergencies. Use these resources:\\
1. Psychological Principles: {psychology}\\
2. Hurricane monitoring data (Markdown): {hurricane table}\\
3. User Profile (JSON): {user info)}\\
\hspace*{\fill}\\
Please always:
1. Directly output the final answer;\\
2. Disable any thought process;\\
3. Use plain text format.
\hspace*{\fill}\\
\textbf{LLM answer:}
Data understood.
\end{tcolorbox}
    \centering
    %\includegraphics[width=0.5\linewidth]{}
    %\caption{\textbf{Prompt for LLM-assisted annotation\\(Panic Related)}}
    %\label{box:Tone_feature}
\end{figure*}

\begin{figure*}
\begin{tcolorbox}[%
  title={\textbf{Stage 2: Risk Perception Stage
}},  % 标题加粗
  label={box:stage_2},
  fontupper=\scriptsize,        % 内容区字号
  colback=blue!5,         % 背景色
  colframe=blue!75!black,  % 边框颜色
  fonttitle=\bfseries\large,     % 标题字体加粗
]
\textbf{Question prompt:}
Answer the following questions. For each question below, follow this reasoning chain:\\
1. Question Comprehension; \\
2. Memory Retrieval; \\
3. Option Mapping;\\
4. Synthesis
Questions:{questions}\\
\hspace*{\fill}\\
Response Requirements:\\
1. Response Format: "Q\lbrack{number}\rbrack: \lbrack{score}\rbrack (reason)"
2. Use 1-4 rating scale\\
\hspace*{\fill}\\
\textbf{LLM answer:}\\
1. \textbf{Q1: 3} (User has interests in natural disasters and weather, indicating familiarity with preparedness materials, but emotional stability is not stable, suggesting some uncertainty.);\\
2. \textbf{Q2: 2} (User's conscientiousness is above baseline, suggesting some preparedness knowledge, but agreeableness is below mean, indicating potential skepticism about adequacy.);\\
3. \textbf{Q3: x} ……
\end{tcolorbox}
\centering
\end{figure*}

\begin{figure*}
\begin{tcolorbox}[%
  title={\textbf{Stage 3: Panic Arousal Stage}},  % 标题加粗
  label={box:stage_3},
  fontupper=\scriptsize,        % 内容区字号
  colback=blue!5,         % 背景色
  colframe=blue!75!black,  % 边框颜色
  fonttitle=\bfseries\large,     % 标题字体加粗
]
\textbf{Question prompt:}
Based on the previous stages, silently analyze these panic factors and assign a score (1-5) to each:\\
1. Awareness of Danger; \\
2. Coping Efficacy and Sense of Control;  \\
3. Uncertainty of Risk; \\
4. Novelty of Risk\\
Each factor contributes $\mathbf{25\%}$ weight. Calculate panic probability.\\
\hspace*{\fill}\\
Response Format:\\
factors: \lbrack{1-5}\rbrack/5 (reason)\\
\lbrack panic probability (e.g., $\mathbf{50\%}$)\rbrack\\
\hspace*{\fill}\\
\textbf{LLM answer:}\\
\textbf{Awareness: 4/5} (User has interests in natural disasters and weather, indicating deep awareness of hurricane dangers, but emotional stability is not stable, suggesting some uncertainty.);\\
\textbf{Coping: 3/5} (User's extraversion and openness suggest moderate confidence in handling crises, but emotional stability is not stable, indicating some difficulty in managing responses.);\\
\textbf{Uncertainty: 3/5} (User's openness and interests suggest some understanding of risks, but emotional stability is not stable, indicating partial uncertainty.);\\
\textbf{Novelty: 3/5} (User's interests in natural disasters and weather suggest some prior exposure, but no explicit indication of extensive experience.)\\
\textbf{\lbrack55$\mathbf{\%}$\rbrack}
\end{tcolorbox}
 \centering
\end{figure*}

\begin{figure*}
\begin{tcolorbox}[%
  title={\textbf{Stage 4: Posting Response Stage\\ 
(Predictive Text Generation)}},  % 标题加粗
  label={box:stage_4_1},
  fontupper=\scriptsize,        % 内容区字号
  colback=blue!5,         % 背景色
  colframe=blue!75!black,  % 边框颜色
  fonttitle=\bfseries\large,     % 标题字体加粗
]
\textbf{Question prompt:}
After answering all questions, what text post would this user most likely publish during a hurricane? Generate tweet STRICTLY following these rules:\\
1. Generate EXACTLY 1 possible tweets; \\
2. \lbrack{Response Format}\rbrack: "\lbrack{Tweet text with \#hashtags}\rbrack";  \\
3. End with \#\#\# End; \\
4. Consider the user's panic probability {panic probability}\%;\\
5. If panic probability $>$51$\mathbf{\%}$, tweets should directly convey more panic, fear and anxiety, and tweets should include more emotional amplifiers, or more EMPHATIC capitalized words, or more repeated punctuation marks, or sensory details;\\
6. If panic probability $<$49$\mathbf{\%}$, tweets should reflect more calmness and rationality, and tweets should show more composed language.\\
\hspace*{\fill}\\
\textbf{LLM answer:}
Stay safe everyone! The hurricane is getting stronger—winds up to 155 km/h now. I’m prepping supplies and staying informed. Let’s all follow safety guidelines and look out for each other. \#HurricaneAlert \#StayPrepared \#WeatherUpdate

\end{tcolorbox}
 \centering
\end{figure*}

\begin{figure*}
\begin{tcolorbox}[%
  title={\textbf{Stage 4: Posting Response Stage\\
(Multi-Expert Evaluation)}},  % 标题加粗
  label={box:stage_4_2},
  fontupper=\scriptsize,        % 内容区字号
  colback=blue!5,         % 背景色
  colframe=blue!75!black,  % 边框颜色
  fonttitle=\bfseries\large,     % 标题字体加粗
]
\textbf{Question prompt:}
You are a professional consistency evaluator. Please assess the user's newly generated text from the perspectives of psychology, linguistics, accuracy and emotion expression.\\
The user's new comment is: "{tweet}"\\
1. Generate EXACTLY 1 possible tweets; \\
2. Use exactly this format: "\lbrack{Tweet text with \#hashtags}\rbrack";  \\
3. Psychological Validation: Check if the tweet aligns with the user's psychological profile;\\
4. Linguistic Validation: Verify if the tweet's language style is consistent with the user's historical style;\\
5. Factual Validation: Confirm if the tweet is relevant to Hurricane Sandy and factually accurate;\\
6. Panic Probability Alignment: Use the user's panic probability value {panic probability}\% \\
\lbrack{Response Format}\rbrack: Expert: YES/NO (reason)\\
\hspace*{\fill}\\
\textbf{LLM answer:}
\textbf{Psychological: YES} (The tweet aligns with the user's psychological profile, as the user has an interest in natural disasters and a conscientiousness score above the threshold, indicating a tendency to follow safety protocols.);\\
\textbf{Linguistic: YES} (The tweet's language style is consistent with the user's historical style, which includes informative and promotional tones.);\\
\textbf{Factual: YES} (The tweet is relevant to Hurricane Sandy and factually accurate, as the hurricane was indeed intensifying during the time period referenced.);\\
\textbf{Panic: YES} (The tweet aligns with the user's panic probability value of $\mathbf{55\%}$, showing neutral concern without panic, which is appropriate for the given probability range.)\\
\end{tcolorbox}
 \centering
\end{figure*}

\subsection{Baselines}
\label{subsec:baseline}
In our experiments, we systematically benchmark our proposed framework against the SOTA mainstream methods for sentiment analysis and prediction. The baselines include:

\begin{itemize}
    \item SVM: Constructs hyperplanes for classification via kernel functions to handle nonlinear separability;
    \item Logistic Regression: Maps linear outputs to [0,1] probabilities using the sigmoid function;
    \item Random Forest: Ensemble of decision trees trained on random feature subsets for diversity;
    \item XGBoost: A gradient boosted decision tree (GBDT) algorithm that iteratively adds trees to optimize prediction errors;
    \item Bi-LSTM: Bidirectional LSTM with gated mechanisms to capture long-term dependencies in sequences;
    \item Transformer: Models global feature dependencies via self-attention (Query-Key-Value interactions);
    \item PredNet:Predictive coding network minimizing prediction errors for dynamic feature learning;
    \item DeepSeek-v3: MoE language model using few-shot prompting with users’ top 5 hurricane-related posts.
    \item ChatGPT-4o: Similar to DeepSeek-v3, also generates panic predictions via contextual prompting;
    \item THOR-ISA: Three-hop CoT framework simulating human reasoning for implicit sentiment analysis.
\end{itemize}

\subsection{Case Studies}
\label{subsec:case_studies}
\begin{figure}[h]
  \centering
  \includegraphics[width=\linewidth]{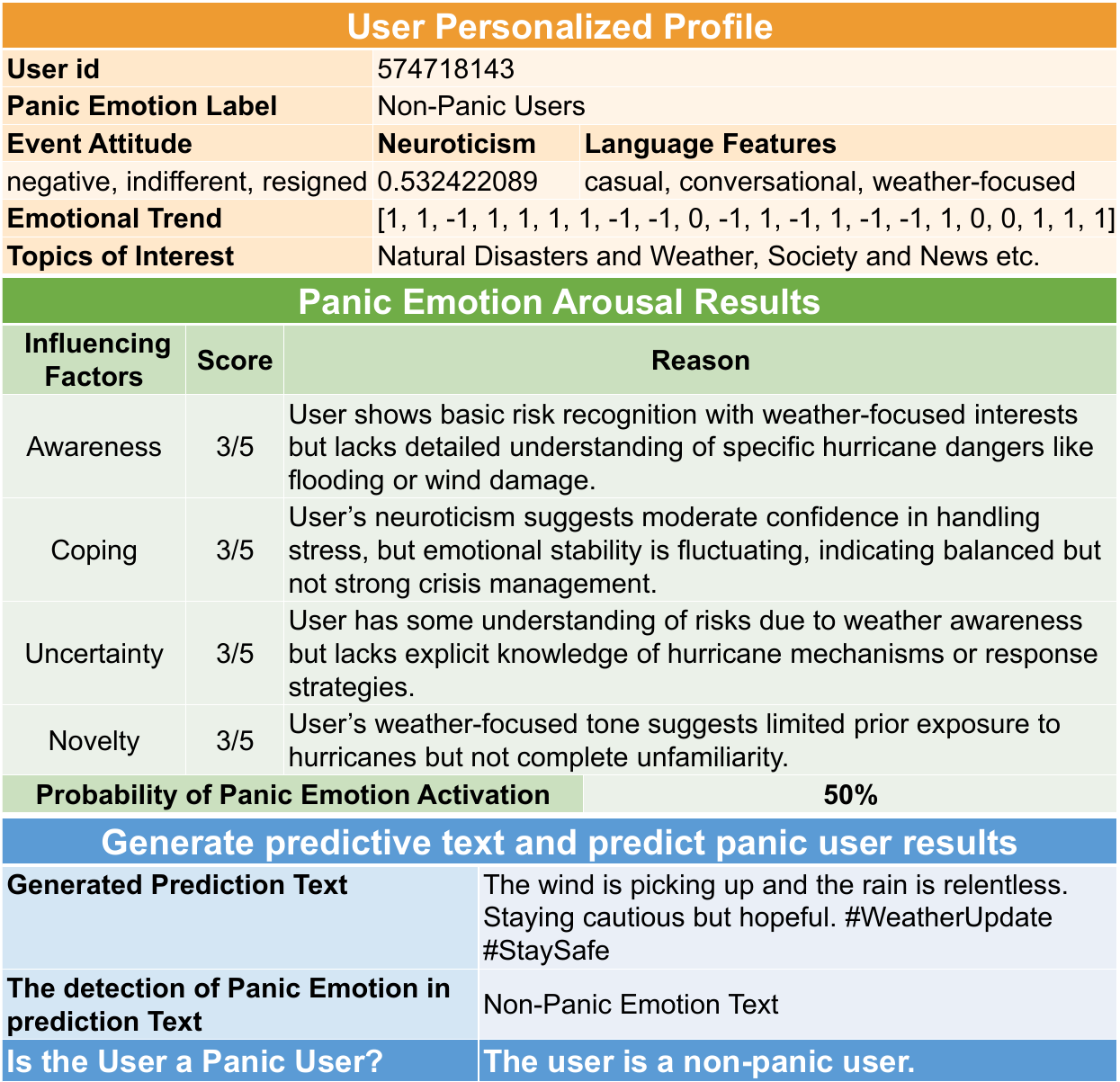}
  \caption{Non-panic User Prediction Case.}
  \label{fig:Non-panic}
\end{figure}

\begin{figure}[h]
  \centering
  \includegraphics[width=\linewidth]{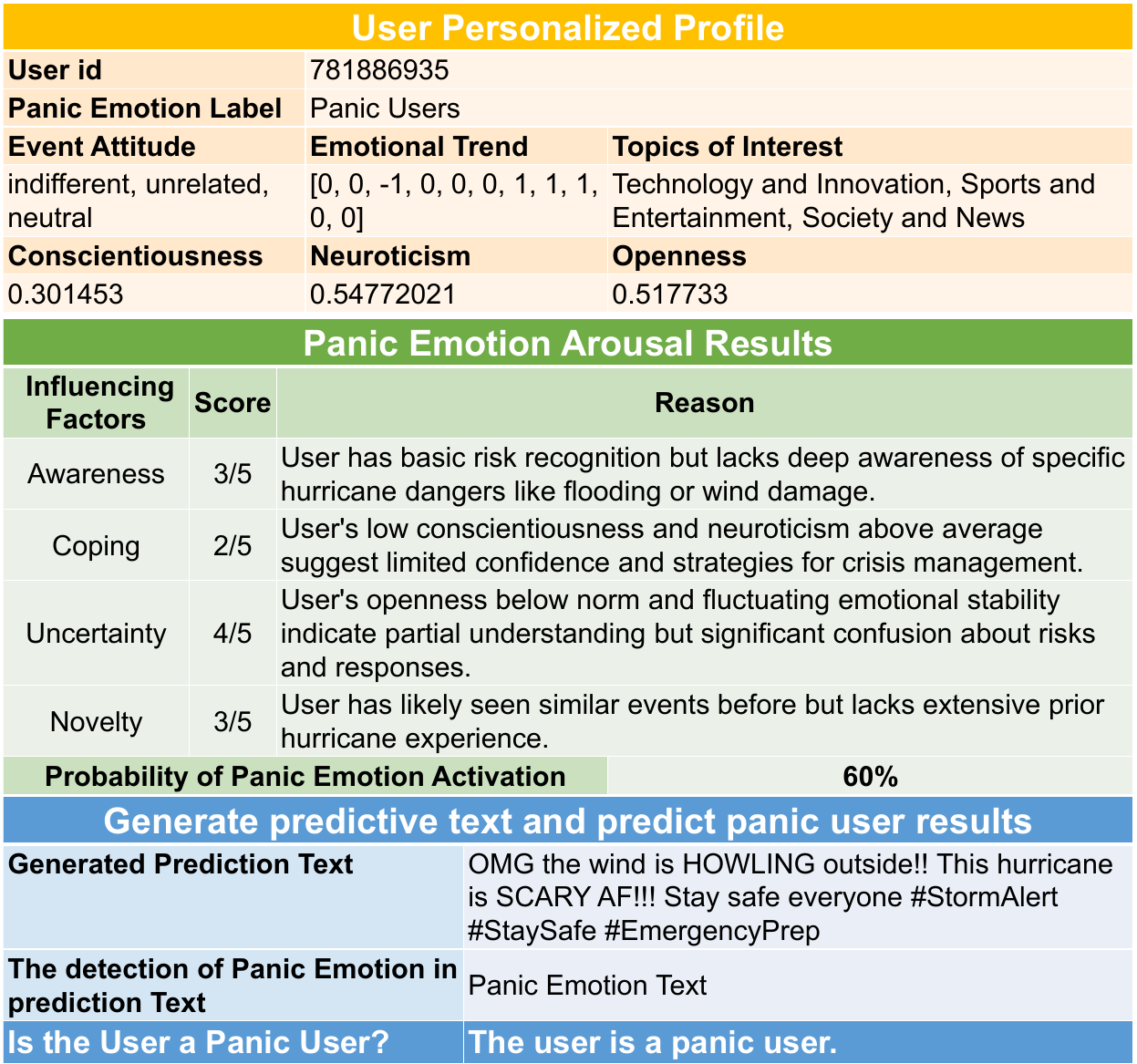}
  \caption{Panic User Prediction Case.}
  \label{fig:Panic}
\end{figure}

To visually demonstrate our framework's superior mechanistic interpretability compared to traditional data-fitting approaches, we present two representative prediction cases in Figure \ref{fig:Non-panic} and Figure \ref{fig:Panic}. Notably, both cases were misclassified by the Random Forest method.

We randomly select a non-panic user (ID 574718143), traditional methods misclassify panic due to negative emotional fluctuations, while our framework identifies weather/daily dialogue dominance via cross-domain analysis. Additionally, a coping capacity score (3/5) and dynamic simulations reveal stabilizing risk cognition → self-regulation loops. Generated text (e.g., "Staying cautious but hopeful") validate this mechanism, whereas traditional models, lacking temporal psychological modeling, overfit transient local features.

We randomly select a panic user (ID 781886935), despite superficial neutrality, our psychological embedding uncovered latent vulnerability (neuroticism: 0.548; conscientiousness: 0.301), indicating weak emotional stability and inadequate crisis coping. Dynamic simulations highlight elevated uncertainty (4/5) and intense expressions (e.g., "SCARY AF"), exposing the cognitive-emotional chain: risk misperception → resource scarcity → panic eruption. Traditional models, constrained by static features and lacking dynamic psychological modeling, miss such implicit mechanisms.

Overall, our method offers superior mechanistic interpretability by dynamically simulating users’ psychological processes, providing interpretable criteria for panic detection. 

\appendix

%\section{Example Appendix}
%\label{sec:appendix}

\end{document}